\documentclass[lettersize,journal]{IEEEtran}
\usepackage{amsmath,amsfonts}
\usepackage{algorithmic}
\usepackage{algorithm}
\usepackage{array}
\usepackage[caption=false,font=normalsize,labelfont=sf,textfont=sf]{subfig}
\usepackage{textcomp}
\usepackage{stfloats}
\usepackage{url}
\usepackage{verbatim}
\usepackage{graphicx}
\usepackage{cite}

\usepackage{pifont}
\usepackage{multirow}
\usepackage{makecell}
\usepackage{enumerate}
\usepackage{amsmath}
\usepackage{amssymb}
\usepackage{booktabs}
\usepackage{color}
\usepackage{framed}
\begin{document}

\title{Text Data-Centric Image Captioning with Interactive Prompts}

\author{
  Yiyu Wang, Hao Luo, Jungang Xu$^*$, Yingfei Sun, Fan Wang
        % <-this % stops a space
  % \thanks{\textit{Corresponding author: Jungang Xu}}
  % \thanks{Yiyu Wang and Yingfei Sun are with the School of Electronic, Electrical and Communication Engineering, University of Chinese Academy of Sciences, Beijing 101408, China; Jungang Xu is with the School of Computer Science and Technology, University of Chinese Academy of Sciences, Beijing 101408, China (e-mail: wangyiyu18@mails.ucas.ac.cn; yfsun@ucas.ac.cn; xujg@ucas.ac.cn).}
  % \thanks{Hao Luo and Fan Wang are with the Alibaba DAMO Academy, Hangzhou 311121, China (e-mail: fan.w@alibaba-inc.com; michuan.lh@alibaba-inc.com).}
}

% \markboth{\LaTeX ~ Template for IEEE Journals}%

% % column for its text to clear the IEEEpubid mark.

\maketitle

\begin{abstract}
  Supervised image captioning approaches have made great progress, but it is challenging to collect high-quality human-annotated image-text data. 
  Recently, large-scale vision and language models (\emph{e.g.}, CLIP) and large-scale generative language models (\emph{e.g.}, GPT-2) have shown strong performances in various tasks, which also provide some new solutions for image captioning with web paired data, unpaired data or even text-only data.
  Among them, the mainstream solution is to project image embeddings into the text embedding space with the assistance of consistent representations between image-text pairs from the CLIP model.
  However, the current methods still face several challenges in adapting to the diversity of data configurations in a unified solution, accurately estimating image-text embedding bias, and correcting unsatisfactory prediction results in the inference stage.
  This paper proposes a new \textbf{T}ext data-centric approach with \textbf{I}nteractive \textbf{P}rompts for image \textbf{Cap}tioning, named TIPCap. 1) We consider four different settings which gradually reduce the dependence on paired data.  2) We construct a mapping module driven by multivariate Gaussian distribution to mitigate the modality gap, which is applicable to the above four different settings. 3) We propose a prompt interaction module that can incorporate optional prompt information before generating captions. Extensive experiments show that our {TIPCap} outperforms other weakly or unsupervised image captioning methods and achieves a new state-of-the-art performance on two widely used {datasets}, \emph{i.e.}, MS-COCO and Flickr30K.
\end{abstract}

  
\newcommand{\etal}{\emph{et al}. }

\section{Introduction}
Image captioning is a typical vision-language task, which aims to automatically generate textual descriptions for given images. In the past few years, although image captioning has made great progress, the models \cite{show_and_tell,show_attend_and_tell,updown,aoanet,Xtransformer,DLCT,PureT,DBLP:journals/tcsv/WuXSYM21,DBLP:journals/tcsv/JiangZH22,DBLP:journals/tcsv/CaoAZW22,DBLP:journals/tcsv/ZhangXDW23} are generally trained on human-annotated image-text data like MS-COCO \cite{mscoco} and Flickr30K \cite{flickr30k}, which are challenging to collect. Some unsupervised works \cite{Feng_etal_unsupervised_IC, DBLP:conf/iccv/Laina0N19} try to mitigate this issue using unpaired image-text data, but still need complex pseudo-training or adversarial training to ensure the semantic alignment between decoder and image.
Recently, the large foundation models such as BERT \cite{BERT}, GPT-2 \cite{GPT2}, T5 \cite{T5}, CLIP \cite{CLIP}, ALIGN \cite{ALIGN} and BLIP \cite{BLIP}, etc, have provide some new solutions for the vision-language tasks, including image captioning. Based on their strong generalization ability and image-text representation alignment, some methods \cite{cc12m,zerocap,magic,capdec,decap,close} use low-cost paired image-text data collected from web (abbreviated as \textit{web data} in this paper) or even text-only data to train models which can even exhibit zero-shot caption capability.

\begin{figure}[t]
  \centering
  \includegraphics[width=\linewidth]{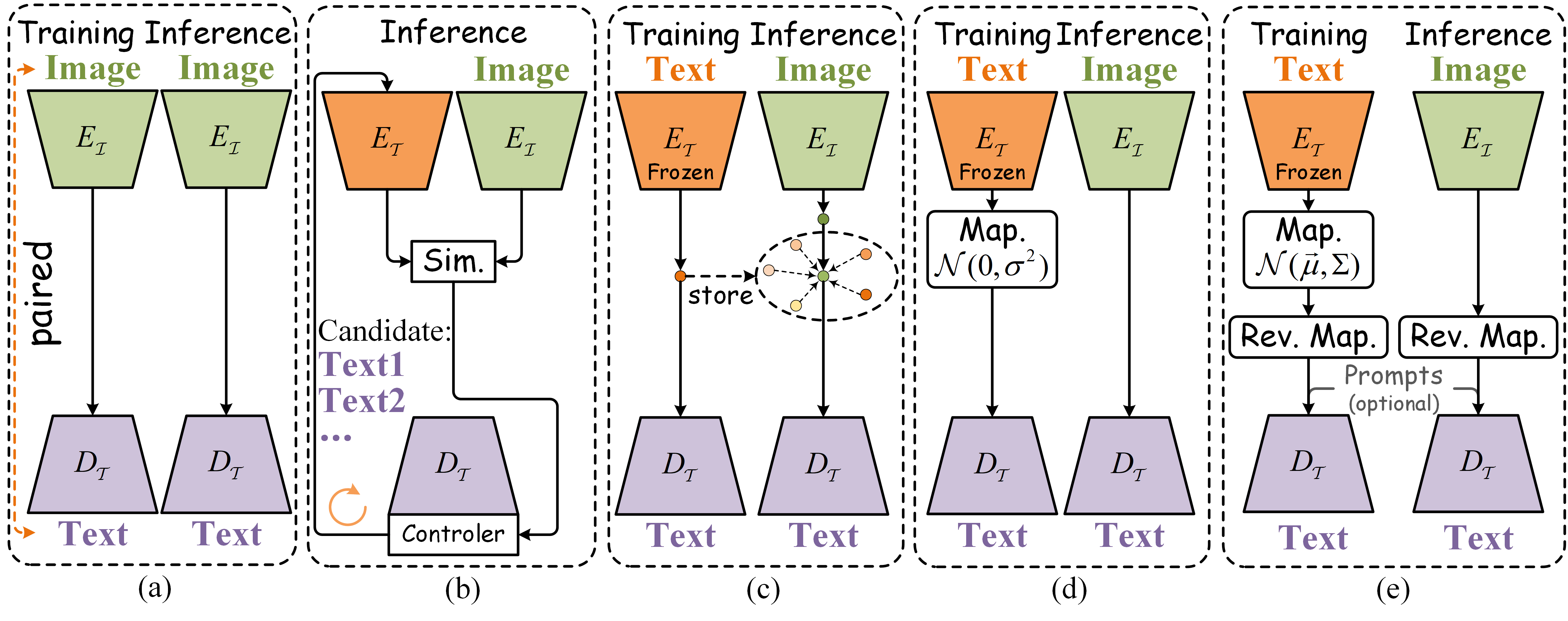}
    \caption{Comparison of different methods, $E_\mathcal{I}$, $E_\mathcal{T}$ and $D_\mathcal{T}$ indicate image encoder, text encoder and text decoder respectively. (a) supervised method. (b) ZeroCap and MAGIC. (c) DeCap. (d) CapDec and CLOSE. (e) our approach.}
    \label{fig:pipeline_intro}
\end{figure}

Specifically, there are two main research lines showing promising potential in image captioning. First, some foundation models (\emph{e.g.} BLIP \cite{BLIP}, BLIP2 \cite{BLIP2}, OFA \cite{OFA}, and SEEM \cite{SEEM}, etc.) trained on large-scale web data successfully unify multiple vision-language understanding and generation tasks, allowing them to directly predict captions for images. However, due to the low-quality labels and significant noise in web data, these models still need adding high-quality image-text datasets such as MS-COCO in the training/fine-tuning data to achieve comparable performance. Second, another line is to leverage the image-text representation alignment capability of the CLIP model to reduce the cost of acquiring training data. It is generally easier to obtain a textual corpus than to obtain high-quality image-text data. Since CLIP can provide consistent feature representations for image-text pairs, some works train caption models with text-only data or little additional paired data.

This paper focuses on the second {research line}, because it is a more caption-specific and resource-friendly solution. Additionally, aligning image-text representations rather than directly predict captions can reduce the requirements of the foundation model on the label quality of image-text data. Among the related works shown in the Fig.~\ref{fig:pipeline_intro}, ZeroCap \cite{zerocap} and MAGIC \cite{magic} generate multiple text and then determining the predicted caption according to the feature similarity between image and text calculated by the CLIP model. However, the untrained generation process and the frequent CLIP text encoder forward limit both the model performance and efficiency. DeCap \cite{decap} proposes another efficient and text-only required method, which builds a support memory using all text features. In the inference stage, the image embedding can be projected into the text embedding space. However, the support memory restricts its ability to scale up to extensive data. Latest CapDec \cite{capdec} and CLOSE \cite{close}, the closest paradigm to ours, make a assumption that the feature bias between a image-text pair in the CLIP embedding space can be estimated with a Gaussian distribution of $\mathcal{N}(0, \sigma^2)$. The value of $\sigma$ is estimated from few paired MS-COCO data in CapDec{,} but {it} is set as a hyper-parameter in CLOSE. In this way, they needs text-only training data because the text embedding can be projected into the image embedding space. 

Although the above methods propose some ingenious solutions, we still point out some crucial issues here. 1) Above methods are usually only compatible with one or two specific data configurations. However, in real-world applications, users have very different data configurations. For example, apart from the text corpus, some web data, few high-quality image-text data like MS-COCO or some web image data can also be provided. How to propose a unified solution to deal with different data configurations? 2) The popular assumption that image-text feature bias is an independent Gaussian distribution may be sub-optimal because correlations exist between different feature dimensions. How to propose a better approximation? 3) These methods inevitably output unsatisfactory results. Can we allow the model to handle user-provided prompts (such as the object in the image) to improve predictions?

Based on the above motivations, {we} proposes a new approach TIPCap that is text data-centric with interactive prompts for image captioning, as shown in Fig.~\ref{fig:pipeline_intro} (e). 
Specifically, our {TIPCap} combines CLIP and GPT-2 to fully leverage the advantages of pre-trained models and contains three extra key modules: a mapping module, a reverse mapping module, and a prompt interaction module.
Firstly, we have taken four different data settings {into account}, which almost cover the vast majority of data configuration scenarios. A unified solution is proposed to estimate text-to-image embedding maps.
Secondly, taking into account the correlation between feature dimensions, the mapping module is driven by multivariate Gaussian distribution instead of independent Gaussian distribution, which aims to mitigate the modality gap by performing a simple projection from CLIP text embedding space to CLIP image embedding space. The reverse mapping module performs a weak projection from CLIP image embedding space back to CLIP text embedding space for stronger robustness. During inference, our TIPCap no longer needs mapping module but directly inputs CLIP image embedding into reverse mapping module and follow-up modules to generate captions.
Thirdly, the prompt interaction module endows TIPCap with the ability to fuse additional prompt information to generate higher-quality descriptions. 
% The prompt tokens will be input into the text decoder to provide prompt information.
With these modules, TIPCap can be trained on text-centric data, and predict captions which can be further improved with manual prompts for a given image.

To evaluate our {approach}, we conduct extensive experiments on two commonly used {datasets}: MS-COCO \cite{mscoco} and Flickr30K \cite{flickr30k}. 
The results demonstrate that our {approach} significantly outperforms existing weakly or unsupervised approaches, and achieves a new state-of-the-art performance.

Our major contributions can be summarized as follows:

\begin{enumerate}[(1)]
  \item We propose a new approach {TIPCap} for image captioning, which provides a unified solution for four settings with different data configurations;
  \item The mapping module utilizes multivariate Gaussian distribution to mitigate the modality gap effectively and outperforms independent Gaussian distribution; our model is able to handle prompt information, which further enhances the flexibility;
  \item Extensive experiments demonstrate the effectiveness of TIPCap and achieve a new state-of-the-art performance.
\end{enumerate}

\section{Related Work}
\subsection{Image Captioning}
\subsubsection{Supervised Approaches}
Inspired by the development of deep learning methods in machine translation \cite{BLEU,DBLP:conf/emnlp/ChoMGBBSB14}, most of existing models utilize encoder-decoder framework. Earlier works \cite{show_and_tell, show_attend_and_tell, show_observe_and_tell, DBLP:conf/cvpr/YouJWFL16} adopt CNN to extract image features and decode them into sentence by LSTM \cite{lstm}. Xu \etal \cite{show_attend_and_tell} introduce an attention mechanism which can dynamically focus on salient regions of the given image.
After that, Anderson \etal \cite{updown} propose to use Faster R-CNN \cite{FasterRCNN} as encoder and achieve significant improvement. {Some} subsequent works \cite{aoanet,DBLP:conf/mm/DongLXX21,DBLP:journals/tcsv/WuXSYM21,DBLP:conf/mm/NieL0LL021,Xtransformer} follow this paradigm.
Recently, transformer-based models have demonstrated excellent performance in image captioning task \cite{ORT,DBLP:journals/tcsv/YuLYH20,Xtransformer,M2Transformer,PureT}.
Although supervised methods have achieved impressive results, high-quality human-annotated paired image-text data is essential.

\subsubsection{Zero-shot Approaches}
Zero-shot image captioning aims to generate description without human-annotated data. While ZeroCap \cite{zerocap} and MAGIC \cite{magic} realize zero-shot image captioning by combining CLIP and GPT-2, and both of them introduce weak visual control cues through the cosine similarity between generated text and given image. Specifically, ZeroCap relies on gradient descent to update the context cache of GPT-2 to make the output match given images; and MAGIC proposes a new decoding strategy to regularize the generated word to be close to given image and previously generated context. However, frequent CLIP text encoder forward slow the inference speed significantly. 
Wang \etal \cite{CLMs} argue that the above methods are prone to fall into harmful contextual language prior and ignore the visual information of given image. 
% Aiming {at} this issue, they propose to utilize text data to fine-tune GPT-2 with the input of CLIP text embedding and extracted nouns. During inference, the input is directly replaced with CLIP image embedding and anchor nouns extracted by object detector. However, they ignore the modality gap issue between CLIP image and text embedding.
% DeCap \cite{decap}, CapDec \cite{capdec} and CLOSE \cite{close} propose zero-shot image captioning methods using text-only dataset. 
DeCap \cite{decap} contains a frozen CLIP and a lightweight text decoder. During training, DeCap takes the CLIP text embedding as input to reconstruct its textual sequence, and stores all training CLIP text embeddings as a support memory. {During} inference, they project the image embedding into CLIP text embedding space by calculating a weighted sum of all embeddings in support memory based on the cosine similarity. 
CapDec \cite{capdec} and CLOSE \cite{close} also perform zero-shot image captioning using text-only data and have a similar paradigm, estimating the modality gap between image and text by an independent Gaussian distribution $\mathcal{N}(0, \sigma^2)$. 

\subsection{Vision-Language Models}
Inspired by BERT \cite{BERT} and its task-agnostic pre-training paradigm, 
a line of works \cite{LXMERT,ViLBERT,UNITER,Oscar} have extended it to Vision-Language (VL) for learning joint representations of image content and natural language, these models directly rely on pre-trained object detector to extract image region features and employ a multimodal encoder to fuse multi-modal features by solving tasks such as masked language modeling (MLM) and image-text matching (ITM).
Another line of works (\emph{e.g.} CLIP \cite{CLIP} and ALIGN \cite{ALIGN}) construct unimodal encoders for image and text separately without densely cross-domain connection, and perform pre-training from scratch on web-scale noisy dataset using only a contrastive loss. 
ALBEF \cite{ALBEF} believes that the learning of image-text interactions and alignment are both important, so introduces a contrastive loss to align image and text before multimodal encoder.
The above models are all encoder-based and are not easy {to} directly transfer to text generation tasks. Considering this issue, BLIP \cite{BLIP} proposes a unified model for VL understanding and generation. Specifically, BLIP constructs unimodal encoders and text decoder for different VL tasks and applies parameter sharing between text encoder and decoder except for the self-attention layers. 
% In this paper, we apply CLIP model as the encoder like existing zero-shot image captioning methods.

\begin{figure*}[htpb]
  \centering
  \includegraphics[width=0.85\linewidth]{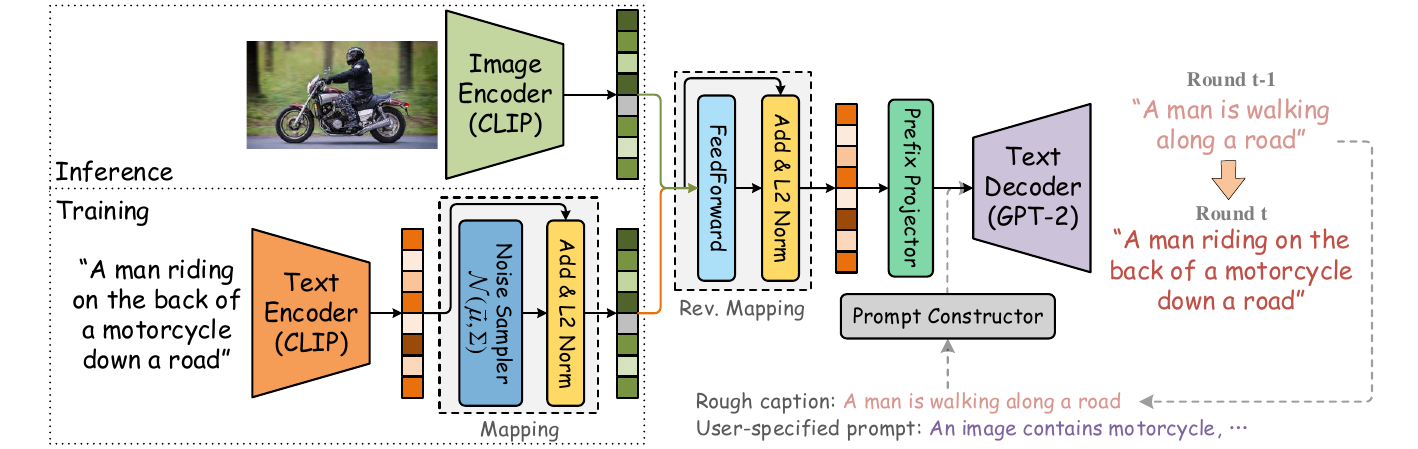}
  \caption{
    {The overall framework of our approach}. Our approach TIPCap is based on a pre-trained CLIP model and a pre-trained GPT-2 model. 
    During training,
    we first exploit CLIP to extract CLIP text embedding and project it into CLIP image embedding space by a mapping module; 
    then we reconstruct text embedding by a reverse mapping module and inject optional prompt information; 
    finally, GPT-2 generates description.
    In the inference stage, we no longer need the mapping module but directly feed CLIP image embedding into reverse mapping module and follow-up modules to generate captions.
  }
  \label{fig:framework}
  % \vspace{-1.em}
\end{figure*}

\section{Method}
The overall framework of our {approach called} TIPCap is shown in Fig.~\ref{fig:framework}. 
We first introduce model architecture, 
and four settings with different data configurations, 
then the details of interactive prompts,
and our training objectives.

\subsection{Model Architecture}
\label{model_architecure}
As shown in Fig.~\ref{fig:framework}, we utilize the frozen CLIP \cite{CLIP} and GPT-2 \cite{GPT2} following existing methods. 
In addition, our TIPCap contains a mapping module, a reverse mapping module, and a prefix projector. 

%, in which the CLIP model is frozen. Overall, we aim to perform training driven by text data instead of human-annotated paired data during training, and the model can directly generate textual description for given image during inference.

Given a text sequence $T$, we first extract the CLIP text embedding $T_e$ and project it into CLIP image embedding space by a mapping module:
\begin{align}
  T_e &= \operatorname{CLIP}_\text{text}(T) \\
  T_e^I &= \operatorname{Mapping}(T_e, \mathcal{N}(\vec{\mu}, \Sigma))
\end{align}
where $\mathcal{N}(\vec{\mu}, \Sigma)$ indicates a multivariate Gaussian distribution with mean $\vec{\mu}$ and covariance $\Sigma$. Briefly, the mapping module performs a simple projection by adding a noise $\epsilon$ obeying $\mathcal{N}(\vec{\mu}, \Sigma)$ into $T_e$.

Then, a reverse mapping module re-projects $T_e^I$ back into CLIP text embedding space:
\begin{align}
  T_e^T &= \operatorname{Reverse-Mapping}(T_e^I)
\end{align}
% where the reverse mapping module consists of two linear layer with ReLU activation function in between (\emph{i.e.} FeedForward \cite{Transformer}). 

The prefix projector module projects CLIP embedding from CLIP dimension to GPT-2 dimension. 
Finally, $P_e$ with optional prompt embedding $Pt_e$ are incorporated as input of GPT-2 to generate captions:
\begin{align}
  P_e &= \operatorname{Prefix-Projector}(T_e^T) \\
%   P_e' &= \operatorname{Prompt-Interaction}(P_e, \{p_i\}) 
%   \label{eq:interactive_prompt} \\
  T' &= \operatorname{GPT-2}(P_e, Pt_e)
\end{align}

The inference forward is slightly different from training forward described above: We no longer need the mapping module but directly extract CLIP image embedding and feed it into reverse mapping module and follow-up modules.
% In addition, the prompt information is optional. $\{p_i\}$ will be replaced with padding tokens when no prompt is provided.
% (see Section~\ref{interactive_prompts} for more details).

\begin{table}
\centering
  \caption{Taking the experiments on MS-COCO as example, a comparison of different data settings.}
  \label{tab:data_settings}
  \begin{tabular}{l|c|c}
    \toprule
    & base data & external data \\
    \hline
    Setting 1 & COCO text & 1\% COCO paired data. \\
    Setting 2 & COCO text & 100K YFCC15M paired data. \\
    Setting 3 & COCO text & 5K YFCC15M image data. \\
    Setting 4 & COCO text & None \\
    \bottomrule
  \end{tabular}
  % \vspace{-1.5em}
\end{table}

\subsection{Estimation of $\vec{\mu}$ and $\Sigma$}
\label{data_settings}
% As described above, the 
The mapping module is driven by a multivariate Gaussian distribution $\mathcal{N}(\vec{\mu}, \Sigma)$ to mitigate the modality gap, which brings up a crucial issue: how to obtain the mean $\vec{\mu}$ and covariance $\Sigma$ that can effectively characterize the true modality bias $\delta$ between CLIP image embedding and text embedding?

In this paper, we consider four settings with different data configurations{, which can cover the majority of real-world scenarios} (text corpus $T_{corpus}$ is available in each setting):

\begin{enumerate}[(1)]
  \item Setting 1: Few human-annotated high-quality paired data $ \langle I_{human},$ $T_{human} \rangle $ is available, where $T_{human}$ is \textbf{homologous} to $T_{corpus}$;
  \item Setting 2: Low-quality paired web data $\langle I_{web}, $ $T_{web} \rangle$ is available, where $T_{web}$ is \textbf{heterologous} to $T_{corpus}$;
  \item Setting 3: No paired data, but a few source-agnostic image data $I_{any}$ is available;
  \item Setting 4: No any extra data, only $T_{corpus}$ is available.
\end{enumerate}
% Now we introduce the details of estimating the mean $\vec{\mu}$ and covariance $\Sigma$ in above four different settings.

\noindent\textbf{Setting 1.}
We first calculate the embedding difference of human annotated paired data:
\begin{align}
  \delta &\simeq I_{e,human} - T_{e,human}
\end{align}
% \begin{align}
%   \delta_{human} &= I_{e,human} - T_{e,human}
% \end{align}
where $I_{e,human}$ and $T_{e,human}$ indicate CLIP image embedding and CLIP text embedding respectively. 

As mentioned before, $\mathcal{N}(\vec{\mu}, \Sigma)$ aims to characterize the modality bias. 
Here $T_{human}$ is paired with $I_{human}$, and also homologous to $T_{corpus}$, so we can directly adopt the mean $\vec{\mu}_{\delta}$ and covariance $\Sigma_{\delta}$ of $\delta$ as a tight estimation of $\vec{\mu}$ and $\Sigma$.

\noindent\textbf{Setting 2.}
Similar to setting 1, we can calculate the embedding difference of web data, and use the mean and covariance of embedding difference as an estimation. But it performs worse, due to $T_{web}$ and $T_{corpus}$ are heterologous. We propose to apply a simple correction to alleviate this issue:
\begin{align}
  \delta &\simeq I_{e,web}-\operatorname{Correct}(T_{e,web}) \notag\\
   &= I_{e,web} - (T_{e,web} + \delta_{w\rightarrow c}) \\
   &= \delta_{web} - \delta_{w\rightarrow c} \notag
\end{align}
where $\delta_{web}$$\sim$$\mathcal{N}(\vec{\mu}_{web}, \Sigma_{web})$ and $\delta_{w\rightarrow c}$$\sim$$\mathcal{N}(\vec{\mu}_{w\rightarrow c}, \Sigma_{w\rightarrow c})$, {subscript} ${}_{w\rightarrow c}$ indicates ``web data to corpus data'', which aims to achieve a domain alignment from $T_{web}$ to $T_{corpus}$.
% $\vec{\mu}_{w\rightarrow c}$ and $\Sigma_{w\rightarrow c}$ are calculated from the embedding difference between $T_{e,corpus}$ and $T_{e,web}$. 
Since there is no pairwise relationship between $T_{corpus}$ and $T_{web}$, $\operatorname{Correct}(\cdot)$ is just a global and rough estimation of the domain alignment.

\noindent\textbf{Setting 3.}
In this setting, we extend the mapping module to be trainable instead of pre-defined parameters. Specifically, the covariance can be denoted as $\Sigma = LL^\top$ by cholesky decomposition, where $L$ is a lower triangular matrix with the same size as $\Sigma$. 
Through reparameterization trick \cite{VAE}, the noise of $\epsilon \sim \mathcal{N}(\vec{\mu}, \Sigma)$ can be re-formulated as follows:
\begin{align}
    \epsilon = L \epsilon' + \vec{\mu}, \quad \epsilon'\sim\mathcal{N}(\vec{0}, I)
\end{align}
Therefore, we can carry out the mean $\vec{\mu}$ and matrix $L$ as trainable parameters. 

The difficulty lies in how to drive the training of $\vec{\mu}$ and $L$ toward the correct direction. For this purpose, we introduce a few source-agnostic image data $I_{any}$, and calculate the embedding difference:
\begin{align}
  \delta \simeq \delta_{any} = I_{e, any} - T_{e, corpus}
\end{align}
where $\delta_{any}\sim\mathcal{N}(\vec{\mu}, \Sigma)$ if $I_{any}$ is paired with $T_{corpus}$. Here we relax this requirement and assume that $\delta_{any}\sim\mathcal{N}(\vec{\mu}, \Sigma)$ is also roughly satisfied even though image and text are unpaired, to derive our training objective. 

Again we apply reparameterization trick, and have:
\begin{align}
  \delta_{any} &\sim \mathcal{N}(\vec{\mu}, \Sigma) \simeq L\epsilon + \vec{\mu}, \quad \epsilon\sim\mathcal{N}(\vec{0}, I)
\end{align}
then we apply a simple transformation to $\delta_{any}$, and result in:
\begin{align}
  \epsilon = L^{-1}(\delta_{any}-\vec{\mu}) \sim \mathcal{N}(\vec{0}, I)
\end{align}

From the above, our training goal is to make $\epsilon$ more close to a standard Gaussian distribution of $\mathcal{N}(\vec{0}, I)$:
\begin{align}
  \mathcal{L}_\text{Map} = KL(L^{-1}(\delta_{any}-\vec{\mu})\|\mathcal{N}(\vec{0}, I))
\end{align}

\noindent\textbf{Setting 4.}
Setting 4 explores a more extreme data configuration, where only text data is available. 
We follow the paradigm similar to setting 3 and apply $\mathcal{L}_\text{Map}$ to optimize trainable parameters $\vec{\mu}$ and $L$.
% As described in Setting 3, the 
The calculation of $\mathcal{L}_\text{Map}$ relies on the modality bias between CLIP image embedding and CLIP text embedding.
However, if we use the bias between $T_e^I$ and $T_e$ to {optimize} the loss $\mathcal{L}_\text{Map}$, the model is prone to falling into trivial solutions (collapse), because the modality bias between the output and input of mapping module, \emph{i.e.} $T_e^I$ and $T_e$, is directly sampled from $\mathcal{N}(\vec{\mu}, \Sigma)$.
To address such issue, we propose several specific designs.

Firstly, we follow other works to apply several asymmetric designs to enhance the robustness and avoid mode collapse. 
Different from the mapping module consisting of an addition operation between the input and a sampled multivariate Gaussian distribution noise, the reverse mapping module is designed as a FeedForward layer, and aimed at re-project the output of mapping module back into CLIP text embedding space.
Then, we {do} not use the original text embedding $T_e$ but the reconstructed one $T_e^T$ to {optimize} $\mathcal{L}_\text{Map}$.
Then, $\mathcal{L}_\text{Map}$ can be calculated as follows:
\begin{gather}
  \mathcal{L}_\text{Map} = KL( L^{-1}(\delta_{pseudo} - \vec{\mu}) \|\mathcal{N}(\vec{0}, I)) \\
  \delta_{pseudo} = T_e^I - T_e^T
\end{gather}
where $T_e^I$ and $T_e^T$ indicate the output of mapping module and reverse mapping module respectively. Since the reverse mapping module does not perform strict reconstruction of CLIP text embedding, which also introduces asymmetry to obtain a more robust performance. In order to ensure the unity of our TIPCap, we also apply the reverse mapping module in other settings except Setting 4.

Secondly, since the lack of real image data to introduce corresponding latent prior information, we employ a relational knowledge distillation loss:
\begin{align}
  \mathcal{L}_\text{Disti} &= KL(\mathcal{S}_{T_e^I}\|\mathcal{S}_{T_e})
\end{align}
where $\mathcal{S}_{T_e^I}$ and $\mathcal{S}_{T_e}$ indicate internal cosine similarity matrix of $T_e^I$ and $T_e$ respectively. Employing $\mathcal{L}_\text{Disti}$ is aiming to encourage $T_e^I$ to have a similar internal cosine similarity to $T_e$, which provides a constraint to ensure that $T_e^I$ and $T_e$ are semantically related and avoids mode collapse.

\subsection{Interactive Prompts}
Inevitably, image captioning models output unsatisfactory sentences, sometimes with factual errors or missing objects. Based on this issue, we hope to endow our model with the ability to deal with additional prompt information to generate information-enhanced captions. 

Inspired by the supervised fine-tuning of InstructGPT \cite{instructGPT}, we construct prompts as textual sentences and serve as a part of input of GPT-2 model, as shown in Fig.~\ref{fig:framework}. One full prompt contains two parts: a rough caption predicted by the model and a user-specified prompt sentence correcting the caption. 
The main difficulty focuses on collecting user-specified prompts sentences during training, which should contain information ignored in rough captions but existed in ground truth captions, to introduce positive information for training guidance. 

% To avoid complex and expensive manual annotation, w
% and explore a simple strategy to generate user-specified prompts

We devide the training into two stages: 1) We perform the first stage of training without introducing prompts and get a base captioning model $model_\text{base}$, which aims to endow our model with the ability to generate rough captions for the second stage training. 2) Then, we initialize $model_\text{prompt}$ with parameters of $model_\text{base}$ to perform the second stage of training with introducing prompts. To avoid complex and expensive manual annotation, we explore a simple strategy to generate user-specified prompts. At each training step, $model_\text{base}$ is frozen to generate rough captions $\langle c_r\rangle$. For user-specified prompt, we do extract nouns or noun phrases set $\{p_i\}_{i=1}^N$ by part-of-speech tagging from ground truth captions $\langle c_{gt}\rangle$. Furthermore, aiming to preserve positive prompt information, we remove nouns or noun phrases that appear in $\langle c_r\rangle$ from $\{p_i\}_{i=1}^N$ to get filtered set $\{p_i'\}_{i=1}^{N'}$. 

Taking Fig.~\ref{fig:framework} as example, for the round $t$ generation, we have $\langle c_r\rangle = \emph{``A man is walking along a road.''}$, $\{p_i'\}_{i=1}^{N'}$ = \emph{\{``motorcycle''\}}, and $\langle c_{gt}\rangle$ = \emph{``A man riding on the back of a motorcycle down a road.''}
The corresponding full prompt sentence $Pt$ is constructed by prompt constructor as follows:
% \vspace{-0.5em}
\begin{framed}
  \vspace{-0.5em}
  \noindent
    \textbf{Reference:} \emph{A man is walking along a road.}\\
    \textbf{Prompt:} \emph{An image contains motorcycle.}\\
    \textbf{Prediction:} \emph{A man riding on the back of a motorcycle down a road.}
  \vspace{-0.5em}
\end{framed}
% \vspace{-0.5em}

Note that we hope our TIPCap keeps the ability to perform general captioning task (\emph{i.e.} generate captions without “reference” and “prompt”), thus prompt information is \textbf{NOT} always essential. 
Furthermore, when the generated caption is good and descriptive enough, it is also not neccessary to introduce prompts and perform the second inference. 

% To achieve the above goal, we add corresponding samples without prompt information during training. Specifically, we replace the top 2-line sub-sentence of the full prompt sentence with padding tokens. See Appendix for more examples of interactive prompts.

\begin{figure}[ht]
  \centering
  \includegraphics[width=0.95\linewidth]{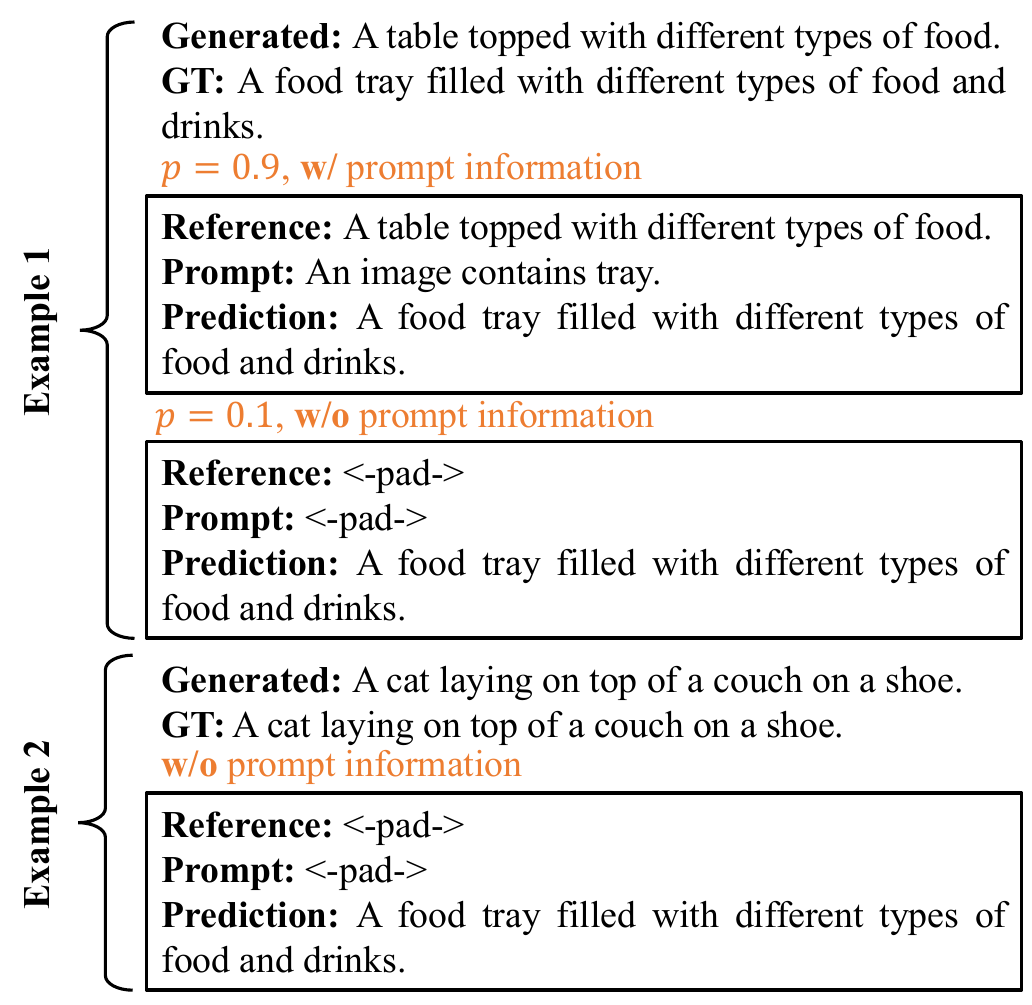}
  \caption{Examples of constructed full prompt sentences during stage 2 training.}
  \label{fig:appendix_examples_prompt_train}
  % \vspace{-1em}
\end{figure}

Based on the above considerations, we replace the prompt sentences with padding tokens with a probability of $p=0.1$ when performing the stage 2 training (refer to ``Implementation details'' in Section~\ref{experimental_setting}); in addition, we also replace the prompts with padding when generated caption is the same as the ground truth caption. Examples of constructed full prompt sentences are shown in Fig.~\ref{fig:appendix_examples_prompt_train}.

% After training based on the above strategy of constructing full prompt sentences, we can perform inference with / without introducing interactive prompts, Fig.~\ref{fig:appendix_examples_prompt} gives more examples of captions generated by TIPCap, where \textbf{``Reference''} and \textbf{``Prediction''} indicate the generated captions without and with simulated prompt (\emph{i.e.} \textbf{``Prompt (simulated by sampling)''}), respectively.

\subsection{Objectives}
\label{objectives}
\noindent\textbf{Language Modeling Loss.}
For a mini-batch of $n$ texts $\{T^i\}_{i=1}^n$, where $T^i=\{T_1^i, ..., T_L^i\}$, we optimize our model by applying Maximum Likelihood Estimation (MLE):
\begin{align}
  \mathcal{L}_\text{MLE} = -\frac{1}{n\times L}\sum_{i=1}^n\sum_{j=1}^{L}\log p_\theta\left(T_j^i|T_{1:{j-1}}^i\right)
\end{align}
where $\theta$ denotes the parameters of our model.

\noindent\textbf{Reverse Mapping Reconstruction Loss.}
For our reverse mapping module, we define $\mathcal{L}_{Recons}$ to constrain the reconstruction relationship of $T_e^T$ to $T_e$:
\begin{align}
  \mathcal{L}_\text{Recons} &= \mathcal{L}_\text{Cosine} + \mathcal{L}_\text{CL}
\end{align}
where $\mathcal{L}_\text{Cosine}$ and $\mathcal{L}_\text{CL}$ denote cosine embedding loss and contrastive loss \cite{CLIP} respectively, and are defined as follows:
\begin{align}
  \mathcal{L}_\text{Cosine} = \frac{1}{n}\sum_{i=1}^n(1-\mathcal{S}_{i,i})&,
  \mathcal{L}_\text{CL} = -\frac{1}{2}(\mathcal{L}_{CL}^{\mathcal{S}} + \mathcal{L}_{CL}^{\mathcal{S}^\top}) \\
  \mathcal{L}_\text{CL}^{\mathcal{S}} = \frac{1}{n}\sum_{i=1}^n\log&\frac{\exp(\mathcal{S}_{i,j}/\tau)}{\sum_{j=1}^n\exp(\mathcal{S}_{i,j}/\tau)}
\end{align}
where $\tau=0.1$, $\mathcal{S}\in\mathbb{R}^{n\times n}$ indicates the cosine similarity matrix between $T_e^T$ and $T_e$. 
Intuitively, $\mathcal{L}_\text{Cosine}$ is introduced for hard reconstruction, and $\mathcal{L}_\text{CL}$ aims to relax the reconstruction constraint. Because we hope that $T_e^T$ has similar semantics to $T_e$ without perfect reconstruction.

\section{Experiments}
\subsection{Experimental Setting}
\label{experimental_setting}
% \noindent\textbf{Datasets.}
\subsubsection{Datasets}
For fair performance comparison, we conduct experiments on both MS-COCO \cite{mscoco} and Flickr30K \cite{flickr30k} datasets, and follow ``Karpathy'' split \cite{DBLP:journals/pami/KarpathyF17}.
Furthermore, YFCC15M \cite{yfcc100m} which has been used to train CLIP is adopt as low-quality paired web data for setting 2 and 3. 

 \textbf{MS-COCO} is a widely used dataset in image captioning task, which consists of 123,287 images, and each is paired with 5 human-annotated captions. 
\textbf{Flickr30K} is another human-annotated dataset similar to MS-COCO, where each image is also paired with 5 reference captions, but only contains 31,000 images. 
\textbf{YFCC15M} is a subset of the large-scale web-collected dataset YFCC100M, where each image is annotated with a weakly paired alt-text. 

Taking experiments on MS-COCO as examples, TABLE~\ref{tab:data_settings} shows a comparison of the available data under four different settings, which cover the majority of real-world scenarios. Specifically, we sample 1\% paired MS-COCO data for setting 1, about 100K paired YFCC data for setting 2, and 5K YFCC image data for setting 3.

% For setting 1, we randomly sample 1\% paired data from MS-COCO and Flickr30K in default. For setting 2, we randomly sample about 100K paired data from YFCC15M. For setting 3, we randomly sample about 5K images from YFCC15M.

% \noindent\textbf{Evaluation metrics.}
\subsubsection{Evaluation metrics}
Following the common paradigm, we adopt five widely used metrics for evaluation, including BLEU-$N$ \cite{BLEU}, METEOR \cite{METEOR}, ROUGE-L \cite{ROUGE}, CIDEr \cite{CIDEr}, and SPICE \cite{SPICE}, which are denoted as B-$N$, M, R, C, and S for simplicity.

% \noindent\textbf{Implementation details.}
\subsubsection{Implementation details}
Our TIPCap model is implemented in PyTorch \cite{pytorch} and trained on 4 Nvidia Tesla V100 (32GB) GPUs. 
We first train our model \textbf{without} introducing interactive prompts for 10 epochs, where the learning rate is warmed-up to $5e^{-4}$ during 1250 steps and decayed linearly;
then we froze parameters of the mapping module and reverse mapping module and train our model \textbf{with} interactive prompts for another 5 epochs with learning rate of $1e^{-6}$, in this training stage. 
For optimization, we set the batch size on each GPU to 32 and adopt AdamW optimizer \cite{adamw} with a weight decay of 0.1 in both above stages and beam size is set to 5 during inference.
Trained models and source code will be released.

\begin{table*}
  % \scriptsize
  \centering
  \caption{In-domain captioning results on MS-COCO and Flickr30K. The superscript $^*$ indicates that results are from MAGIC \cite{magic}. 
  For weakly or unsupervised approaches, they use CLIP with different backbone as Encoder. DeCap \cite{decap} constructs a lightweight Transformer Decoder (T.D.) instead of pre-trained GPT-2. CLOSE \cite{close} uses T5 model \cite{T5}. Setting 1-4 are abbreviated as S1-4.}
  \label{tab:performance_to_sota}
  \begin{tabular}{l|c|c|cccccc|cccccc}
    \toprule
    \multirow{2}{*}{Method} & \multirow{2}{*}{Encoder} & \multirow{2}{*}{Decoder} & \multicolumn{6}{c|}{MS-COCO} & \multicolumn{6}{c}{Flickr30k} \\
    \cline{4-9}  \cline{10-15}
    & & & B-1 & B-4 & M & R & C & S & B-1 & B-4 & M & R & C & S\\
    \hline
    \textbf{\emph{Fully Supervised:}} & & & & & & & & & & & & \\
    % \hline
    BUTD    & & & 77.2 & 36.2 & 27.0 & 56.4 & 113.5 & 20.3 & - & 27.3 & 21.7 & -    & 56.6 & 16.0 \\
    UniVLP  & & & -    & 36.5 & 28.4 & -    & 116.9 & 21.2 & - & 30.1 & 23.0 & -    & 67.4 & 17.0 \\
    ClipCap & & & -    & 33.5 & 27.5 & -    & 113.1 & 21.1 & - & -    & -    & -    & -    & -    \\
    Oscar   & & & -    & 36.5 & 30.3 & -    & 123.7 & 23.1 & - & -    & -    & -    & -    & -    \\
    \hline
    \textbf{\emph{Weakly or Unsupervised:}} & & & & & & & & & & & & \\
    % \hline
    ZeroCap$^*$ & ViT-B/32 & GPT-2 & 49.8 &  7.0 & 15.4 & 31.8 & 34.5 &  9.2 & 44.7 &  5.4 & 11.8 & 27.3 & 16.8 &  6.2 \\
    MAGIC       & ViT-B/32 & GPT-2 & 56.8 & 12.9 & 17.4 & 39.9 & 49.3 & 11.3 & 44.5 &  6.4 & 13.1 & 31.6 & 20.4 &  7.1 \\
    DeCap       & ViT-B/32 & T.D. & -    & 24.7 & 25.0 & -    & 91.2 & 18.7 & -    & 21.2 & 21.8 & -    & 56.7 & 15.2 \\
    CapDec      & RN50x4 & GPT-2 & 69.2 & 26.4 & 25.1 & 51.8 & 91.8 & -    & 55.5 & 17.7 & 20.0 & 43.9 & 39.1 & -    \\
    CLOSE       & ViT-L/14 & T5 & -    & 22.1 & 23.7 & -    & 81.2 & 17.7 & -    & -    & -    & -    & -    & -    \\
    % CLOSE (Mult.)       & -    & 29.5 & 25.7 & -    & 97.8 & 18.3 & -    & -    & -    & -    & -    & -    \\
    CLOSE w/Tuned Noise      & ViT-L/14 & T5 & -    & 28.6 & 25.2 & -    & 95.4 & 18.1 & -    & -    & -    & -    & -    & -    \\
    % CLOSE (Mult.)$^\dag$       & -    & 29.5 & 25.6 & -    & 98.4 & 18.3 & -    & -    & -    & -    & -    & -    \\
    \hline
    TIPCap (S1) & RN50x4 & GPT-2 & 74.6 & 30.7 & 26.7 & 54.2 & 106.7 & 20.3 & 71.1 & 25.6 & 22.5 & 49.1 & 63.7 & 16.2\\
    % \quad\quad \textcolor{gray}{+ 1 prompt} & \\
    \hline
    TIPCap (S2) & RN50x4 & GPT-2 & 72.7 & 28.6 & 25.6 & 52.5 & 100.6 & 19.6 & 68.0 & 23.7 & 21.3 & 47.3 & 57.8 & 15.2 \\
    \hline
    TIPCap (S3) & RN50x4 & GPT-2 & 73.0 & 30.4 & 26.5 & 53.8 & 104.5 & 20.0 & 68.4 & 24.2 & 21.9 & 48.1 & 61.2 & 16.1 \\
    \hline
    TIPCap (S4) & RN50x4 & GPT-2 & 71.3 & 29.8 & 26.2 & 53.4 & 102.1 & 19.4 & 67.5 & 24.0 & 21.7 & 47.7 & 59.4 & 15.9 \\
    TIPCap (S4) & ViT-L/14 & GPT-2 & 73.3 & 31.4 & 26.9 & 54.2 & 106.6 & 20.2 & 69.6 & 26.1 & 23.0 & 49.3 & 65.7 & 17.0 \\
    \bottomrule
  \end{tabular}
\end{table*}

\begin{table*}
    % \scriptsize
    \centering
    \caption{Cross-Domain captioning results. X $\Longrightarrow$ Y means {that the model is trained on dataset X but evaluated on dataset Y.} }
    \label{tab:performance_cross_domain}
    \begin{tabular}{l|cccccc|cccccc}
      \toprule
      \multirow{2}{*}{Method} & \multicolumn{6}{c|}{Flickr30k $\Longrightarrow$ MS-COCO} & \multicolumn{6}{c}{MS-COCO $\Longrightarrow$ Flickr30k} \\
      \cline{2-7}  \cline{8-13}
      & B-1 & B-4 & M & R & C & S & B-1 & B-4 & M & R & C & S\\
      \hline
      MAGIC    & 41.4 &  5.2 & 12.5 & 30.7 & 18.3 &  5.7 & 46.4 &  6.2 & 12.2 & 31.3 & 17.5 & 5.9  \\
      CapDec   & 43.3 &  9.2 & 16.3 & 36.7 & 27.3 & -    & 60.2 & 17.3 & 18.6 & 42.7 & 35.7 & -    \\
      DeCap    & -    & 12.1 & 18.0 & -    & 44.4 & 10.9 & -    & 16.3 & 17.9 & -    & 35.7 & 11.1 \\
      \hline
      TIPCap (S1) & 59.8 & 16.7 & 19.4 & 42.3 & 56.0 & 12.5 & 66.9 & 19.8 & 19.8 & 45.3 & 48.2 & 13.7 \\
      TIPCap (S2) & 55.9 & 14.5 & 17.8 & 40.4 & 47.8 & 11.4 & 63.5 & 17.3 & 18.4 & 43.2 & 41.6 & 12.3 \\
      TIPCap (S3) & 55.9 & 14.2 & 18.4 & 40.6 & 48.7 & 11.9 & 63.7 & 18.6 & 19.2 & 44.0 & 42.8 & 13.0 \\
      TIPCap (S4) & 55.8 & 14.3 & 18.4 & 40.5 & 48.5 & 11.9 & 63.8 & 18.7 & 19.2 & 44.1 & 42.4 & 12.9 \\
      \bottomrule
    \end{tabular}
\end{table*}

\subsection{Comparison with {Existing} Models}
% \noindent\textbf{In-domain captioning.}
\subsubsection{In-domain captioning}
TABLE~\ref{tab:performance_to_sota} shows the in-domain captioning results on MS-COCO and Flickr30K. 
We compare our proposed TIPCap with several approaches with different supervision levels. 
1) Fully supervised approaches, which rely on human-annotated paired data for model training, including BUTD \cite{updown}, UniVLP \cite{uniVLP}, ClipCap \cite{clipcap}, Oscar \cite{Oscar}; 
and 2) weakly or unsupervised approaches, which adopt pre-trained foundation models (\emph{e.g.} CLIP \cite{CLIP}, GPT-2 \cite{GPT2} and T5 \cite{T5}) to perform image captioning on unpaired or text-only data, including ZeroCap \cite{zerocap}, MAGIC \cite{magic}, DeCap \cite{decap}, CapDec \cite{capdec} and CLOSE \cite{close}. Our proposed TIPCap belongs to the second category, weakly or unsupervised approaches.

As shown in TABLE~\ref{tab:performance_to_sota}, we report the performance results of four different data settings.
As expected, TIPCap (Setting 1) performs better than the other three settings, as it estimates the distribution parameter of mapping module from high-quality paired data, which introduces strong and credible prior information of modality bias and can be regarded as an upper bound of our proposed approach. 
TIPCap (Setting 2) uses weakly paired web data but performs worse, because the alt-text is low-quality and has a large margin with training text corpus.
TIPCap (Setting 3) and TIPCap (Setting 4) have less data available but also achieve superior performances that significantly outperforms other state-of-the-art models, which shows the effectiveness of our trainable mapping module. Especially, our TIPCap uses CLIP with RN50x4 backbone and GPT-2 model but outperforms the strong competitor CLOSE that adopts CLIP with ViT-L/14 backbone and T5 model.

Overall, our proposed TIPCap achieves state-of-the-art performance and shows substantial performance improvement compared with recent weakly or unsupervised methods, demonstrating the effectiveness and advantages.

% Unless otherwise specified, all results reported uses RN50x4 CLIP model.

\subsubsection{Cross-domain captioning}
As shown in TABLE~\ref{tab:performance_cross_domain}, we conduct cross-domain experiments to further explore the generalization ability of our approach. Specifically, we train our TIPCap on the source dataset (\emph{e.g.} MS-COCO), but perform inference on a different target dataset (\emph{e.g.} Flickr30K). We compare TIPCap with several text-only methods, including MAGIC \cite{magic}, CapDec \cite{capdec}, and DeCap \cite{decap}. Our TIPCap still outperforms all compared methods, demonstrating the superiority of our {approach} in generalization ability.

\begin{figure*}
  \centering
    \includegraphics[width=\linewidth]{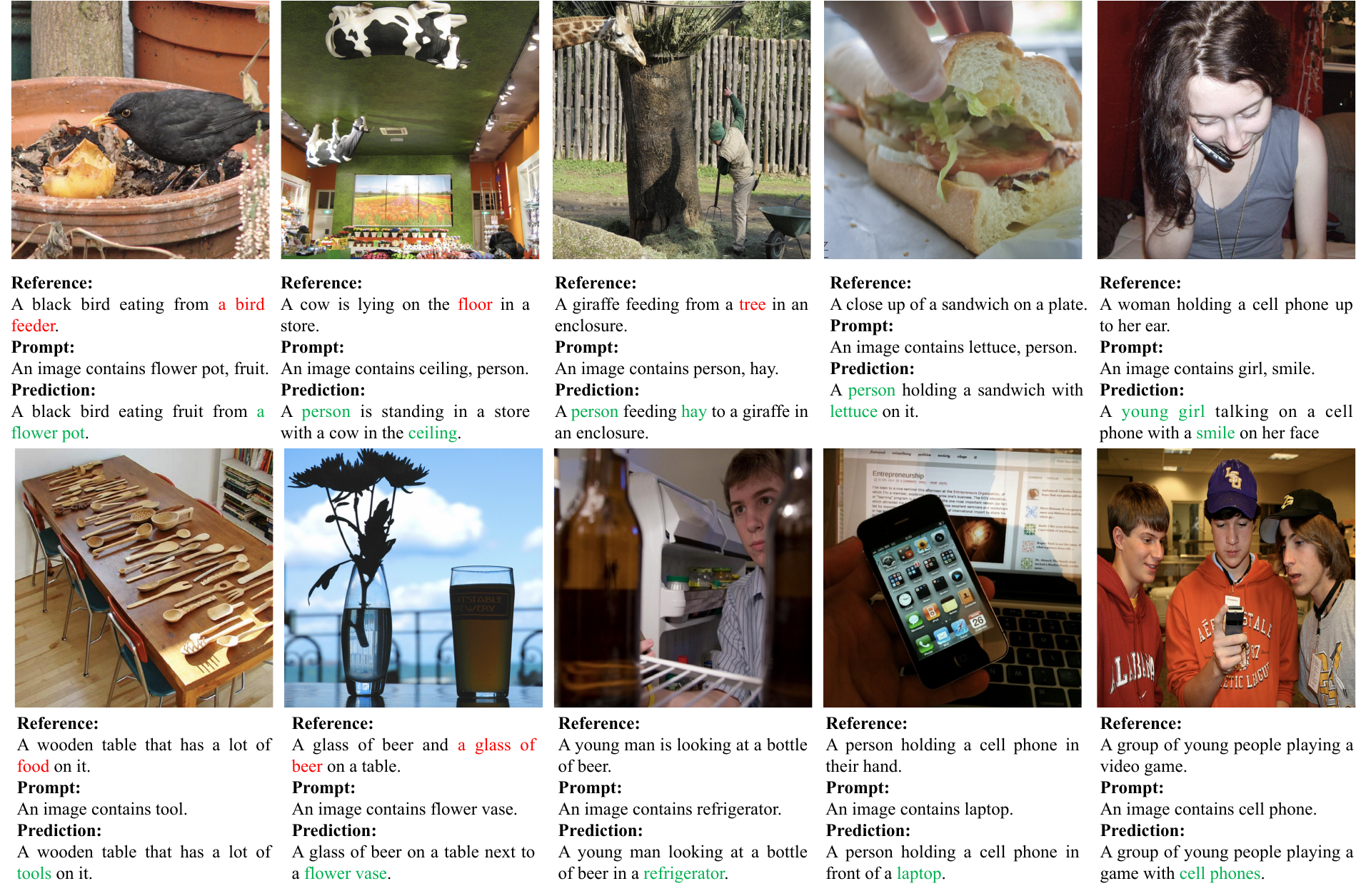}
    \caption{Examples of captions generated by TIPCap with simulated interactive prompts, images come from MS-COCO karpathy test split. \textbf{``Reference''} indicates the generated caption withou prompt information; \textbf{``Prompt''} indicates the simulated user-specified prompt information; \textbf{``Prediction''} shows the new generated caption with prompt information. 
    }
    \label{fig:visualization_examples}
\end{figure*}

\begin{table*}
  \centering
  % \scriptsize
  % \tabcolsep=.6pt
  % \renewcommand\arraystretch{0.8}
  \caption{Performance comparison with simulated interactive prompt information in TIPCap.}
  \label{tab:interactive_prompts}
  \begin{tabular}{c|c|cccccc}
    \toprule
    & & B-1 & B-4 & M & R & C & S \\
    \hline
    \multirow{4}{*}{\rotatebox{0}{\makecell{CLIP\\RN50x4}}}
    & S1 & 77.27$_{\pm.15}$ & 32.73$_{\pm.06}$ & 27.57$_{\pm.06}$ & 55.53$_{\pm.12}$ & 113.17$_{\pm.35}$ & 21.97$_{\pm.06}$\\
    & S2 & 75.93$_{\pm.06}$ & 30.77$_{\pm.12}$ & 26.70$_{\pm.10}$ & 54.03$_{\pm.06}$ & 107.93$_{\pm.06}$ & 21.60$_{\pm.00}$ \\
    & S3 & 78.23$_{\pm.06}$ & 33.90$_{\pm.10}$ & 28.50$_{\pm.00}$ & 56.33$_{\pm.06}$ & 116.27$_{\pm.06}$ & 24.27$_{\pm.06}$\\
    & S4 & 77.80$_{\pm.10}$ & 33.60$_{\pm.17}$ & 28.30$_{\pm.00}$ & 56.20$_{\pm.00}$ & 115.33$_{\pm.06}$ & 23.73$_{\pm.06}$ \\
    \hline
    \multirow{4}{*}{\rotatebox{0}{\makecell{CLIP\\ViT-L/14}}}
    & S1 & 77.53$_{\pm.06}$ & 33.03$_{\pm.12}$ & 28.00$_{\pm.10}$ & 55.80$_{\pm.00}$ & 115.87$_{\pm.23}$ & 22.40$_{\pm.00}$\\
    & S2 & 76.30$_{\pm.10}$ & 31.30$_{\pm.20}$ & 27.03$_{\pm.06}$ & 54.23$_{\pm.06}$ & 110.23$_{\pm.42}$ & 22.03$_{\pm.06}$ \\
    & S3 & 78.57$_{\pm.15}$ & 34.40$_{\pm.20}$ & 28.40$_{\pm.00}$ & 56.27$_{\pm.06}$ & 116.87$_{\pm.21}$ & 24.10$_{\pm.00}$ \\
    & S4 & 78.27$_{\pm.15}$ & 33.90$_{\pm.17}$ & 28.37$_{\pm.06}$ & 56.07$_{\pm.06}$ & 116.57$_{\pm.42}$ & 24.13$_{\pm.06}$ \\
    \bottomrule
  \end{tabular}
  % \vspace{-1em}
\end{table*}

\begin{figure*}
  \centering
  \includegraphics[width=\linewidth]{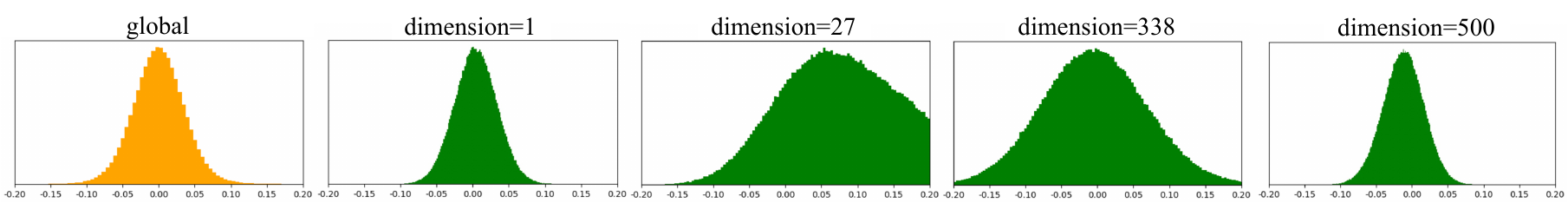}
  \caption{Histogram visualization of the CLIP image and text embedding difference of MS-COCO training set, where orange and green indicate the histogram statistics on all dimensions (global) and specific dimensions (local) separately.}
  \label{fig:vis_hist}
  \vspace{1.em}

  \centering
  \includegraphics[width=\linewidth]{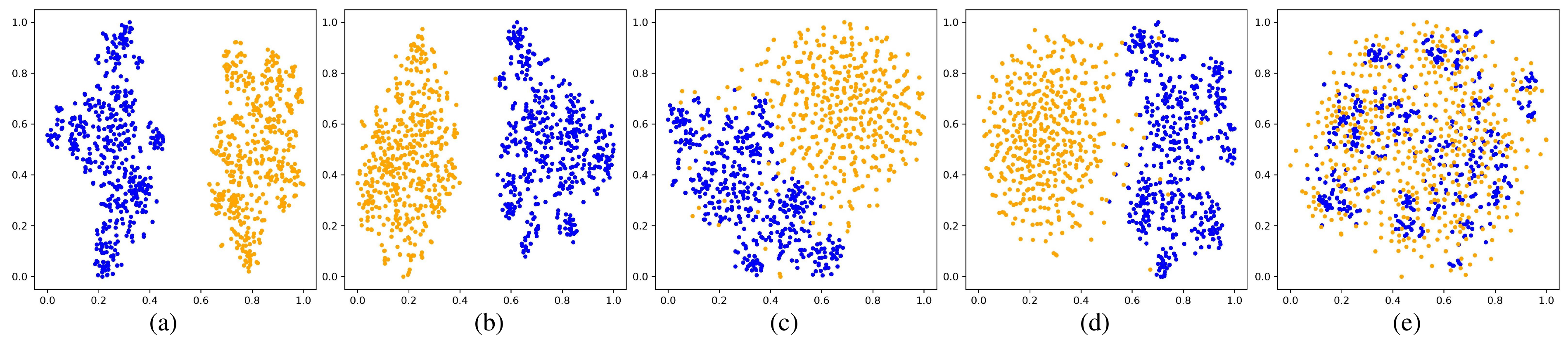}
  \caption{t-SNE visualization of CLIP image and text embeddings from MS-COCO training set, where blue and yellow points indicate image and text embeddings, respectively. 
  (a) without mapping module; 
  (b) mapping module driven by univariate Gaussian distribution $\mathcal{N}(\mu, \sigma^2)$, where $\mu\simeq 0.0009$ and $\sigma\simeq 0.0440$ are estimated from MS-COCO paired data; 
  (c) mapping module driven by $\mathcal{N}(\mu, \sigma^2)$, where $\mu=0$ and $\sigma=\sqrt{0.016}$, which is adopted in CapDec \cite{capdec};
  (d) mapping module driven by $\mathcal{N}(\mu, \sigma^2)$, where $\mu=0$ and $\sigma=0.08$, which is adopted in CLOSE \cite{close};
  (e) mapping module driven by multi-variate Gaussian distribution $\mathcal{N}(\vec{\mu}, \Sigma)$, where $\vec{\mu}$ and $\Sigma$ are estimated from coco paired data.
  }
  \label{fig:appendix_tsne}
\end{figure*}

\begin{table}[h]
  \centering%
  % \scriptsize
  \tabcolsep=2.pt
  \caption{Ablation study of mapping module and reverse mapping module. 
  (A) Our full TIPCap model; 
  (B) TIPCap with the mapping module driven by independent Gaussian distribution; 
  (C) TIPCap model with reverse mapping module removed.
  ``N/A'' refers to ``Not Applicable''.}
  \label{tab:ablation_map_and_rev_map}
  \begin{tabular}{l|cccccc|cccccc}
    \toprule
    &\multicolumn{6}{c|}{CLIP RN50x4} & \multicolumn{6}{c}{CLIP ViT-L/14} \\
    \cline{2-7}  \cline{8-13}
    &B-1 & B-4 & M & R & C & S & B-1 & B-4 & M & R & C & S \\
    \midrule
    \multicolumn{13}{c}{(A) \textbf{TIPCap, $\mathcal{N}(\vec{\mu}, \Sigma)$}} \\
    \midrule
    S1 & 74.6 & 30.7 & 26.7 & 54.2 & 106.7 & 20.3 & 75.4 & 31.2 & 27.2 & 54.5 & 109.7 & 20.9 \\
    S2 & 72.7 & 28.6 & 25.6 & 52.5 & 100.6 & 19.6 & 73.4 & 29.1 & 26.0 & 52.7 & 103.5 & 20.2 \\
    S3 & 73.0 & 30.4 & 26.5 & 53.8 & 104.5 & 20.0 & 73.7 & 31.2 & 26.7 & 53.9 & 107.5 & 20.5 \\
    S4 & 71.3 & 29.8 & 26.2 & 53.4 & 102.1 & 19.4 & 73.3 & 31.4 & 26.9 & 54.2 & 106.6 & 20.2 \\
    \midrule
    \multicolumn{13}{c}{(B) \textbf{TIPCap, $\mathcal{N}(\mu, \sigma^2)$}} \\
    \midrule
    S1 & 68.5 & 24.8 & 24.1 & 49.7 & 89.5 & 18.6 & 66.7 & 22.5 & 23.2 & 46.8 & 87.3 & 18.1 \\
    S2 & 69.4 & 25.5 & 24.4 & 50.2 & 91.6 & 18.7 & 68.0 & 23.5 & 23.5 & 47.7 & 90.1 & 18.3 \\
    S3 & 69.9 & 26.2 & 24.6 & 50.7 & 92.9 & 18.9 & 70.3 & 25.3 & 24.7 & 50.2 & 94.1 & 19.0 \\
    S4 & 70.0 & 25.9 & 24.1 & 50.5 & 91.5 & 18.5 & 69.9 & 25.3 & 24.4 & 49.9 & 93.8 & 18.8 \\
    \midrule
    \multicolumn{13}{c}{(C) \textbf{TIPCap, $\mathcal{N}(\vec{\mu}, \Sigma)$}, \textbf{w/o reverse mapping module}} \\
    % \multicolumn{13}{c}{\textbf{w/o reverse mapping module}} \\
    \midrule
    S1 & 74.1 & 30.2 & 26.2 & 53.7 & 105.0 & 19.8 & 75.3 & 31.3 & 26.9 & 54.4 & 109.2 & 20.8 \\
    S2 & 72.4 & 28.3 & 25.1 & 52.0 & 99.2 & 19.1 & 73.7 & 29.2 & 25.9 & 52.8 & 103.1 & 19.8 \\
    S3 & 71.4 & 30.0 & 26.3 & 53.4 & 102.4 & 19.4 & 73.2 & 30.8 & 26.6 & 53.9 & 105.5 & 20.1 \\
    S4 & \multicolumn{6}{c|}{N/A} & \multicolumn{6}{c}{N/A} \\
    \bottomrule
  \end{tabular}
\end{table}

\subsection{Ablation Studies}
\subsubsection{Impact of Interactive Prompts}
Due to the dynamic nature of the prompt information and the lack of relevant benchmarks, it is not easy to give accurate quantitative results when introducing interactive prompts. 

To verify the effectiveness of interactive prompts, we perform the evaluation on MS-COCO with simulated user-specified prompt information. Specifically, we first apply TIPCap to generate a caption without prompt information as the rough caption, then we do second-time inference to introduce the prompt information, in which the simulated user-specified prompt information is sampled from the nouns or noun key phrases (ignored by rough caption) extracted from corresponding ground truth caption.
We perform three times evaluations in each setting and report the average performance with the standard deviation, as shown in TABLE~\ref{tab:interactive_prompts}. The results demonstrate that the introduction of prompt information brings positive influences on performance. 

Furthermore, in order to make a more intuitive qualitative comparison, we give some examples of captions generated by TIPCap when introducing simulated interactive prompts, as shown in Fig.~\ref{fig:visualization_examples}. In which, 
\textbf{``Reference''} caption is generated by the same TIPCap model without prompt information (\emph{i.e.} rough caption);
\textbf{``Prompt''} indicates the simulated user-specified prompt information constructed by sampling from the ignored ground truth nouns or phrases;
\textbf{``Prediction''} gives the new caption generated by our TIPCap.

\subsubsection{Impact of Mapping Module}
This paper makes an assumption that the feature bias between image-text paired CLIP embeddings can be estimated as a multivariate Gaussian distribution, which is a natural extension of the assumption adopted in CapDec \cite{capdec} and CLOSE \cite{close} that use independent Gaussian distribution.

In this part, we conduct experiments to investigate the effect of the above two strategies. The results are reported in TABLE~\ref{tab:ablation_map_and_rev_map} (A) and (B), we can see that the performances are better when using multivariate Gaussian distribution. 
The results of setting 1 when using $\mathcal{N}(\mu, \sigma^2)$ perform worst and intuitively show that the independent Gaussian distribution is a suboptimal strategy, which can not effectively characterize the feature bias.
Moreover, the results of setting 3 and setting 4 in TABLE~\ref{tab:ablation_map_and_rev_map} (B) demonstrate that our TIPCap is also effective when using independent Gaussian distribution and can achieves competitive performances.

Fig.~\ref{fig:vis_hist} shows simple histogram visualization of the CLIP image and text embedding difference of MS-COCO training set, it can be seen that the distribution of embedding differences in different dimensions are not consistent. 
And it is why we apply multivariate Gaussian distribution instead of independent Gaussian distribution.

As shown in Fig.~\ref{fig:appendix_tsne}, we randomly sample 500 paired image-text data from MS-COCO training set and visualize the CLIP image and text embeddings.
Fig.~\ref{fig:appendix_tsne} (a) shows the clear modality gap between CLIP image embeddings and CLIP text embeddings.
Fig.~\ref{fig:appendix_tsne} (b), (c) and (d) show the influence of mapping module driven by univariate Gaussian distribution.
Fig.~\ref{fig:appendix_tsne} (b) indicates that the modality gap still exists after applying the mapping module driven by $\mathcal{N}(\mu, \sigma^2)$, even if mean $\mu$ and standard deviation $\sigma$ are estimated from MS-COCO paired data. 
CapDec \cite{capdec} and CLOSE \cite{close} also use $\mathcal{N}(\mu, \sigma^2)$ but have a larger standard deviation as shown in Fig.~\ref{fig:appendix_tsne} (c) and (d), which mitigate the modality gap but not significantly.
Fig.~\ref{fig:appendix_tsne} (e) shows that our mapping module driven by multi-variate Gaussian distribution can effectively reduce the modality gap.

\begin{table}[t]
  \centering
  % \scriptsize
  \tabcolsep=2.pt
  \caption{Performance comparison with different prefix length $L$ and different layer number of $N$ in the prefix projector module. Experiments are conducted under setting 4.}
  \label{tab:ablation_prefix_length}
  \begin{tabular}{c|c|cccccc|cccccc}
    \toprule
    & & \multicolumn{6}{c|}{CLIP RN50x4} & \multicolumn{6}{c}{CLIP ViT-L/14} \\
    \cline{3-8}  \cline{9-14}
    & & B-1 & B-4 & M & R & C & S & B-1 & B-4 & M & R & C & S \\
    \hline
    \multirow{6}{*}{L}
    & 2  & 71.1 & 29.3 & 26.0 & 53.2 & 101.1 & 19.2 & 74.0 & 31.6 & 26.5 & 54.2 & 106.7 & 20.0 \\
    & 4 & \textbf{71.3} & \textbf{29.8} & \textbf{26.2} & \textbf{53.4} & \textbf{102.1} & \textbf{19.4} & \textbf{73.3} & \textbf{31.4} & \textbf{26.9} & \textbf{54.2} & \textbf{106.6} & \textbf{20.2} \\
    & 5  & 71.1 & 29.5 & 26.3 & 53.3 & 101.7 & 19.5 & 73.3 & 31.3 & 26.8 & 54.2 & 106.7 & 20.3 \\
    & 10 & 71.0 & 29.3 & 26.0 & 53.1 & 100.5 & 19.2 & 72.0 & 30.7 & 27.3 & 54.4 & 106.5 & 20.7 \\
    & 20 & 71.1 & 29.5 & 26.1 & 53.1 & 101.5 & 19.4 & 73.4 & 30.9 & 26.7 & 53.9 & 106.7 & 20.2 \\
    & 40 & 70.1 & 29.2 & 26.2 & 53.0 &  99.9 & 19.1 & 72.3 & 30.4 & 27.1 & 54.3 & 106.3 & 20.5 \\
    \midrule
    % \hline
    \multirow{6}{*}{N}
    & 1 & 71.2 & 29.9 & 26.1 & 53.2 & 101.5 & 19.1 & 72.9 & 31.1 & 26.5 & 53.7 & 105.2 & 19.9 \\
    & 2 & 71.7 & 30.1 & 26.3 & 53.5 & 101.9 & 19.4 & 73.2 & 31.2 & 26.7 & 54.1 & 106.0 & 20.1 \\
    & 3 &\textbf{71.3} & \textbf{29.8} & \textbf{26.2} & \textbf{53.4} & \textbf{102.1} & \textbf{19.4} & \textbf{73.3} & \textbf{31.4} & \textbf{26.9} & \textbf{54.2} & \textbf{106.6} & \textbf{20.2} \\
    & 4 & 71.6 & 29.9 & 26.2 & 53.4 & 102.1 & 19.3 & 73.8 & 31.1 & 26.7 & 54.1 & 107.0 & 20.5 \\
    & 5 & 71.6 & 30.0 & 26.2 & 53.5 & 102.3 & 19.5 & 74.1 & 31.4 & 26.8 & 54.1 & 108.2 & 20.6 \\
    & 6 & 71.2 & 29.8 & 26.3 & 53.5 & 102.4 & 19.6 & 73.7 & 31.5 & 27.0 & 54.1 & 108.5 & 20.8 \\
    \bottomrule
  \end{tabular}

  \vspace{1em}

  \centering
  % \scriptsize
  \tabcolsep=2.0pt
  \caption{
  Performance comparison with different ratio of paired data (\emph{i.e.} MS-COCO) used in setting 1. }
  \label{tab:ablation_setting1}
  \begin{tabular}{c|cccccc|cccccc}
    \toprule
    & \multicolumn{6}{c|}{CLIP RN50x4} & \multicolumn{6}{c}{CLIP ViT-L/14} \\
    \cline{2-7}  \cline{8-13}
    r & B-1 & B-4 & M & R & C & S & B-1 & B-4 & M & R & C & S \\
    \hline
    0.2\% & 73.3 & 29.4 & 26.1 & 53.4 & 103.0 & 19.7 & 74.7 & 30.3 & 26.7 & 53.9 & 107.5 & 20.6 \\
    0.5\% & 74.2 & 30.5 & 26.6 & 54.1 & 105.8 & 20.1 & 75.3 & 31.2 & 27.1 & 54.7 & 109.2 & 20.9 \\
    1\%   & \textbf{74.6} & \textbf{30.7} & \textbf{26.7} & \textbf{54.2} & \textbf{106.7} & \textbf{20.3} & \textbf{75.4} & \textbf{31.2} & \textbf{27.2} & \textbf{54.5} & \textbf{109.7} & \textbf{20.9} \\
    5\%   & 74.4 & 30.7 & 26.7 & 54.3 & 106.5 & 20.3 & 75.5 & 31.5 & 27.3 & 54.8 & 109.9 & 20.9 \\ 
    20\%  & 74.5 & 31.0 & 26.7 & 54.4 & 107.1 & 20.3 & 75.7 & 31.9 & 27.3 & 55.0 & 110.5 & 21.0 \\
    50\%  & 74.6 & 30.9 & 26.8 & 54.4 & 107.3 & 20.3 & 75.6 & 31.8 & 27.5 & 55.0 & 111.0 & 21.1 \\
    100\% & 74.4 & 30.8 & 26.7 & 54.4 & 106.6 & 20.3 & 75.8 & 31.7 & 27.3 & 54.9 & 110.6 & 21.0 \\
    \bottomrule
  \end{tabular}
\end{table}

\subsubsection{Impact of Reverse Mapping Module}
To explore the effectiveness of our proposed reverse mapping module, we report the performances of our TIPCap with the reverse mapping module removed, as shown in TABLE~\ref{tab:ablation_map_and_rev_map} (C). 
Comparing TABLE~\ref{tab:ablation_map_and_rev_map} (A) and (C), it can be see that the reverse mapping module brings slight performance improvement over all metrics. More importantly, the reverse mapping module is essential for our approaches in setting 4, which makes our proposed TIPCap still applicable when only text data is available. 

\subsubsection{Impact of prefix projector}
% The prefix projector aims to project CLIP embedding from CLIP dimension to GPT-2 dimension, and then input the projected embeddings into GPT-2 as prefix embeddings.
The prefix projector aims to convert CLIP embedding to prefix embeddings as the input of GPT-2 model.
Following ClipCap \cite{clipcap} and CapDec \cite{capdec}, we use a transformer-based architecture as prefix projector.
To explore the effect of prefix projector, we perform ablation studies on prefix length of $L$ and layer number of $N$ respectively, as shown in Table~\ref{tab:ablation_prefix_length}. The bold results denote our default implementation.
% The results are shown in Table~\ref{tab:ablation_prefix_length} and Table~\ref{tab:ablation_prefix_layer}.

From the results, TIPCap shows good robustness when increasing $L$ or $N$. 
{It brings an advantage that we do} not need a heavy module to perform the indispensable projection from CLIP embedding space to GPT-2 embedding space.

\begin{table}[t]
  \centering
  % \footnotesize
  % \scriptsize
    \tabcolsep=2.0pt
    \caption{
    Performance comparison with different number of images used in setting 3. All images are randomly sampled from YFCC15M.}
    \label{tab:ablation_setting3_image_cnt}
    \begin{tabular}{c|cccccc|cccccc}
      \toprule
      & \multicolumn{6}{c|}{CLIP RN50x4} & \multicolumn{6}{c}{CLIP ViT-L/14} \\
      \cline{2-7}  \cline{8-13}
      \#img & B-1 & B-4 & M & R & C & S & B-1 & B-4 & M & R & C & S \\
      \hline
      500 & 73.0 & 30.3 & 26.3 & 53.8 & 104.0 & 19.9 & 73.8 & 31.1 & 26.7 & 54.0 & 107.3 & 20.4 \\
      1K  & 72.9 & 30.3 & 26.3 & 53.8 & 104.2 & 19.9 & 73.9 & 31.2 & 26.7 & 54.0 & 107.3 & 20.4 \\
      5K  & \textbf{73.0} & \textbf{30.4} & \textbf{26.5} & \textbf{53.8} & \textbf{104.5} & \textbf{20.0} & \textbf{73.7} & \textbf{31.2} & \textbf{26.7} & \textbf{53.9} & \textbf{107.5} & \textbf{20.5} \\
      10K & 72.9 & 30.4 & 26.4 & 53.7 & 104.0 & 19.9 & 73.8 & 31.2 & 26.7 & 53.9 & 107.3 & 20.4 \\
      \bottomrule
    \end{tabular}

    \vspace{1em}

    \centering
    % \scriptsize
    \tabcolsep=1.5pt
    \caption{Performance comparison of three different sampled data under setting 1. Experiments are conducted on MS-COCO.}
    \label{tab:appendix_sampled_s1}
    \begin{tabular}{l|cccccc|cccccc}
      \toprule
      & \multicolumn{6}{c|}{CLIP RN50x4} & \multicolumn{6}{c}{CLIP ViT-L/14} \\
      \cline{2-7}  \cline{8-13}
      & B-1 & B-4 & M & R & C & S & B-1 & B-4 & M & R & C & S \\
      \hline
      sample 1 & 
      74.6 & 30.7 & 26.7 & 54.2 & 106.7 & 20.3 & 75.4 & 31.2 & 27.2 & 54.5 & 109.7 & 20.9 \\
      sample 2 & 
      74.8 & 31.0 & 26.7 & 54.3 & 107.2 & 20.5 & 75.3 & 31.4 & 27.1 & 54.5 & 109.9 & 20.8 \\
      sample 3 & 
      74.7 & 30.9 & 26.7 & 54.3 & 107.2 & 20.3 & 75.4 & 31.6 & 27.3 & 54.7 & 110.7 & 21.0\\
      \bottomrule
    \end{tabular}

    \vspace{1em}
    \centering
    \tabcolsep=1.pt
    \caption{Performance comparison of different image source (``YFCC15M'' v.s. ``MS-COCO'') under setting 3. Experiments are conducted on MS-COCO.}
    \label{tab:appendix_data_source_setting3}
    \begin{tabular}{l|cccccc|cccccc}
      \toprule
      & \multicolumn{6}{c|}{CLIP RN50x4} & \multicolumn{6}{c}{CLIP ViT-L/14} \\
      \cline{2-7}  \cline{8-13}
      & B-1 & B-4 & M & R & C & S & B-1 & B-4 & M & R & C & S \\
      \hline
      YFCC15M & 
      73.0 & 30.4 & 26.5 & 53.8 & 104.5 & 20.0 & 73.7 & 31.2 & 26.7 & 53.9 & 107.5 & 20.5 \\
      MS-COCO & 
      72.9 & 30.6 & 26.5 & 53.8 & 104.9 & 20.1 & 74.1 & 31.5 & 26.7 & 54.1 & 107.9 & 20.4 \\
      \bottomrule
    \end{tabular}
\end{table}

\subsubsection{Impact of available external data}
In setting 1, the parameter $\vec{\mu}$ and $\Sigma$ are estimated from human-annotated high-quality paired data. 
We perform estimation using different ratios of paired data to study its effect as shown in Table~\ref{tab:ablation_setting1}.
Theoretically, the estimated parameters can more accurately characterize the modality bias when more paired data is available. 
From the results, we can see that the performance tends to stabilize when more than 1\% paired data is available, which indicates that we can obtain sufficiently accurate estimated parameters without all data.

In setting 3, we use few source-agnostic image data to introduce the latent prior information of CLIP image embedding space. 
To explore its effect, we sample different numbers of image data from YFCC15M dataset for training. The results are reported in Table~\ref{tab:ablation_setting3_image_cnt}. 
When using only 500 images, the performance is still advantageous enough, which indicates the data efficiency of our TIPCap.

\subsubsection{Impact of different sampled data under setting 1}
Under Setting 1, the parameters of our mapping module (\emph{i.e.} mean and covariance) are estimated from sampled paired data (\emph{e.g.} MS-COCO or Flickr30K), which brings a question: whether different sampled data will influence the performance.
To verify the influence of different sampling data, we conduct experiments on MS-COCO dataset. We sample 1\% paired MS-COCO data for the estimation of mean and covariance, and perform three times sampling, resulting in three different sets of parameters. Then we train our TIPCap with these three different estimations of $\mathcal{N}(\vec{\mu}, \Sigma)$.

The results are reported in Table~\ref{tab:appendix_sampled_s1}, and show that the performances are robust to different sampling data, which further demonstrates the effectiveness and robustness of applying multivariate Gaussian distribution to mitigate the modality gap.

\subsubsection{Impact of image source under setting 3}
As described in Section 4.1, we use 5K YFCC images for our TIPCap training under setting 3, which aims to prove that heterologous images are also applicable and effective. 
We also conduct experiments using homologous images on MS-COCO dataset, the results are shown in Table~\ref{tab:appendix_data_source_setting3}. 
Specifically, we adopt the 5K images of MS-COCO karpathy validation split, which are homologous to the training corpus but are not paired and do not affect the performance evaluation on karpathy test split.

From the results, both YFCC images and MS-COCO images achieve comparable performance under setting 3, which indicates that the requirement for image data is source-agnostic.
Combining the results shown in Table 8, it indicates that although available images are few and heterologous with training text corpus, TIPCap can still achieve competitive performance under setting 3.
Compared under setting 4, TIPCap under setting 3 can achieve better performance because we introduce prior information of real image for training. 
Although it is difficult to collect high-quality paired image data for text corpus in real-world scenarios, the image data-efficiency under setting 3 makes the collection cost of image data affordable.

\begin{figure*}[t]
  \centering
  \includegraphics[width=\linewidth]{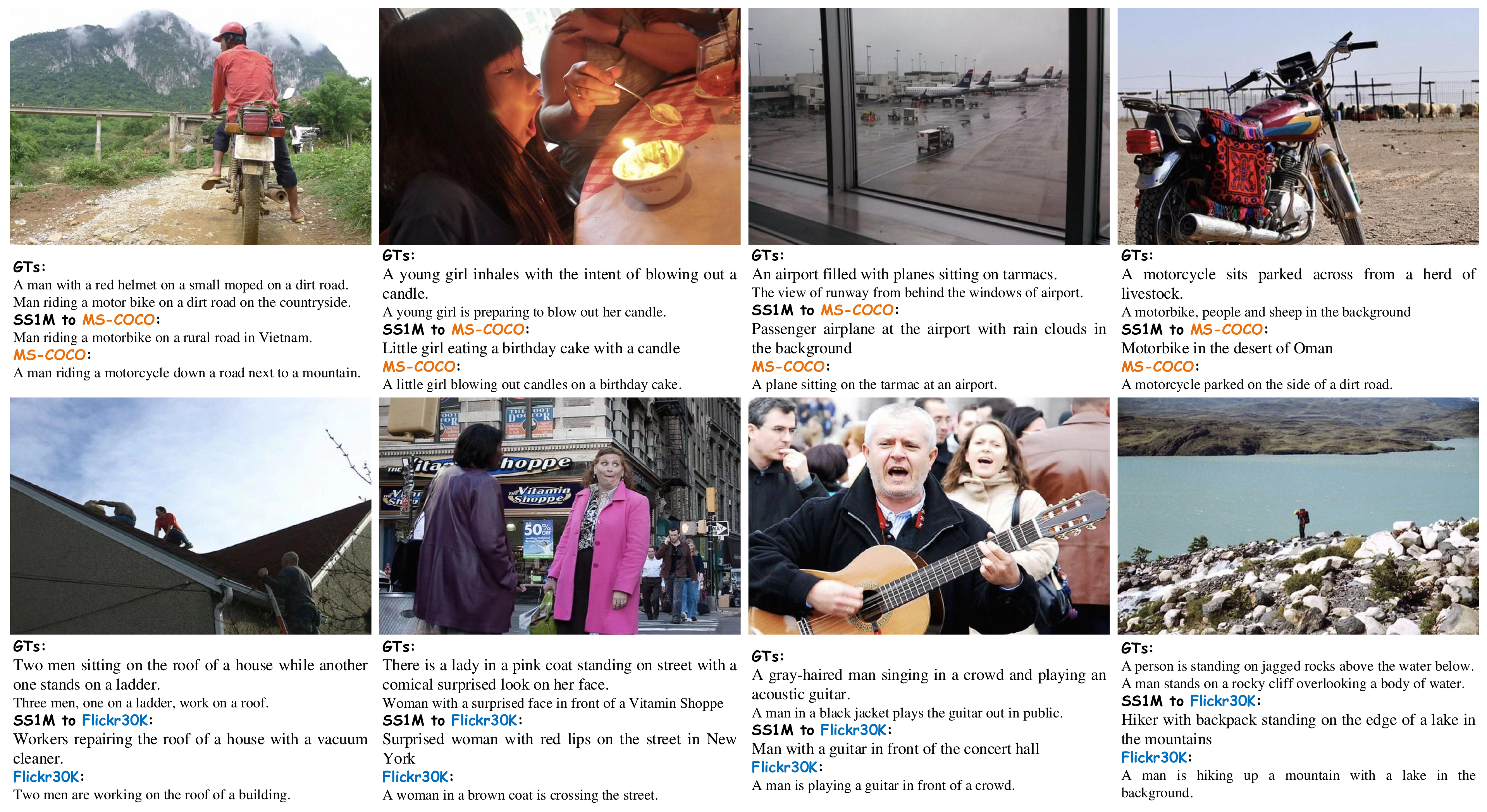}
  \caption{Examples of captions generated by TIPCap. 
  The first two images comes from MS-COCO dataset and the second two images comes from Flickr30K dataset.
  ``SS1M to MS-COCO / Flickr30K'' indicate that captions is generated by TIPCap trained on SS1M. ``MS-COCO'' and ``Flickr30K'' denote TIPCap trained on MS-COCO and Flickr30K, respectively.
  }
  \label{fig:appendix_examples}
  % \vspace{-1em}
\end{figure*}

\begin{table*}
  \centering
  \caption{Comparison of text between SS1M and MS-COCO. 
  }
  \label{fig:appendix_ss1m_vs_coco_text}
  \includegraphics[width=0.7\linewidth]{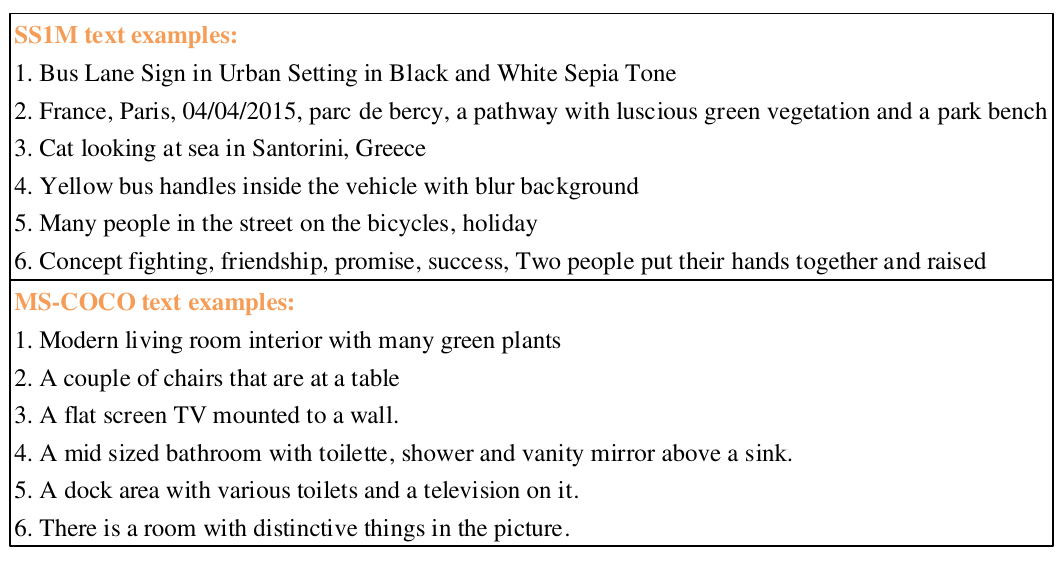}
\end{table*}

\begin{table*}[t]
  \centering
    % \scriptsize
    \caption{Results on MS-COCO karpathy test split and NoCaps validation split, where ``in'', ``near'' and ``out'' indicate \emph{``in-domain''}, \emph{``near-domain''} and \emph{``out-of-domain''} respectively.All models are trained on MS-COCO text corpus.}
    \label{tab:appendix_settings_nocaps}
    \begin{tabular}{l|cccccc|cccc}
      \toprule
      & \multicolumn{6}{c|}{MS-COCO} & \multicolumn{4}{c}{NoCaps val (CIDEr)} \\
      \cline{2-7}  \cline{8-11}
       & B-1 & B-4 & M & R & C & S & in & near & out & overall \\
      \hline
      DeCap & - & 24.7 & 25.0 & - & 91.2 & 18.7 & 65.2 & 47.8 & 25.8 & 45.9 \\
      \hline
      \textbf{CLIP RN50x4:} & & & & & & \\
      TIPCap (S1) & 74.6 & 30.7 & 26.7 & 54.2 & 106.7 & 20.3 & 77.6 & 63.3 & 44.8 & 61.6 \\
      TIPCap (S2) & 72.7 & 28.6 & 25.6 & 52.5 & 100.6 & 19.6 & 71.1 & 56.6 & 39.0 & 55.1 \\
      TIPCap (S3) & 73.0 & 30.4 & 26.5 & 53.8 & 104.5 & 20.0 & 74.6 & 59.2 & 37.1 & 56.9 \\
      TIPCap (S4) & 71.3 & 29.8 & 26.2 & 53.4 & 102.1 & 19.4 & 74.3 & 59.2 & 36.4 & 56.7 \\
      \hline
      \textbf{CLIP ViT-L/14:} & & & & & & \\ 
      TIPCap (S1) & 75.4 & 31.2 & 27.2 & 54.5 & 109.7 & 20.9 & 81.7 & 67.5 & 49.8 & 65.9 \\
      TIPCap (S2) & 73.4 & 29.1 & 26.0 & 52.7 & 103.5 & 20.2 & 76.1 & 61.3 & 44.5 & 60.0 \\
      TIPCap (S3) & 73.7 & 31.2 & 26.7 & 53.9 & 107.5 & 20.5 & 77.3 & 62.7 & 40.6 & 60.9 \\
      TIPCap (S4) & 73.3 & 31.4 & 26.9 & 54.2 & 106.6 & 20.2 & 80.2 & 62.3 & 39.6 & 60.3 \\
      \bottomrule
    \end{tabular}
\end{table*}

\begin{table*}
  \centering
  % \scriptsize
  \caption{Cross-domain image captioning performance. We train our TIPCap using SS1M dataset under setting 3 and setting 1, and perform evaluation on MS-COCO and Flickr30K datasets.}
  \label{tab:appendix_ss1m_s3}
  \begin{tabular}{l|cccccc|cccccc}
    \toprule
    & \multicolumn{6}{c|}{SS1M $\Longrightarrow$ MS-COCO} & \multicolumn{6}{c}{SS1M $\Longrightarrow$ Flickr30K} \\
    \cline{2-7}  \cline{8-13}
     & B-1 & B-4 & M & R & C & S & B-1 & B-4 & M & R & C & S \\
    \hline
    DeCap & - & 8.9 & 17.5 & - & 50.6 & 13.1 & - & - & - & - & - & - \\
    \hline
    TIPCap (S3, CLIP RN50x4) & 54.7 & 11.5 & 18.0 & 35.7 & 53.0 & 13.9 & 52.2 & 9.1 & 15.4 & 32.7 & 32.2 & 11.0 \\
    TIPCap (S3, CLIP ViT-L/14) & 56.5 & 12.3 & 18.6 & 37.2 & 57.3 & 14.5 & 56.6 & 10.7 & 16.2 & 35.3 & 36.3 & 11.3 \\
    \midrule
    TIPCap (S1, CLIP RN50x4) & 62.9 & 16.5 & 20.6 & 41.3 & 69.2 & 16.3 & 59.2 & 13.0 & 17.1 & 37.2 & 38.7 & 11.8 \\
    TIPCap (S1, CLIP ViT-L/14) & 65.1 & 18.3 & 21.5 & 43.2 & 73.9 & 17.0 & 62.6 & 14.4 & 18.1 & 39.3 & 44.0 & 12.5 \\
    \bottomrule
  \end{tabular}
\end{table*}

\subsection{More Generalization Analysis}
To further quantitatively evaluate the generalization performance of our proposed TIPCap, we conduct experiments on two new datasets: NoCaps \cite{nocaps} and SS1M \cite{Feng_etal_unsupervised_IC}.
\subsubsection{Performance on NoCaps dataset} 
MS-COCO is limited to 80 classes, NoCaps \cite{nocaps} provides a benchmark to measure the open-set capability for novel objects (unseen classes in MS-COCO dataset). The NoCaps dataset contains only validation and test sets, and is divided into three parts: 
\emph{in-domain} contains images portraying only MS-COCO classes;
\emph{near-domain} contains both MS-COCO and novel classes;
\emph{out-of-domain} contains only novel classes. Following DeCap \cite{decap}, we evaluate our TIPCap using the validation set on official evaluation server \footnote{https://eval.ai/web/challenges/challenge-page/355/overview}.
% \href{https://eval.ai/web/challenges/challenge-page/355/overview}{official evaluation server}.
Table~\ref{tab:appendix_settings_nocaps} shows the performance comparison on MS-COCO karpathy test split and NoCaps validation split. 
Compared to DeCap, TIPCap achieves significant improvement on all metrics. 

\subsubsection{Performance using SS1M dataset}
SS1M \cite{Feng_etal_unsupervised_IC} is a web-collected text corpus, which uses the name of 80 MS-COCO classes as keywords to crawl image descriptions from Shutterstock
% \href{https://www.shutterstock.com}{Shutterstock}
\footnote{https://www.shutterstock.com}
, resulting in 2,322,628 distinct image descriptions in total that is larger than MS-COCO and Flickr30K.
% In this paper, we follow DeCap to reuse SS1M corpus and remove sentences with more than fifteen words.
Fig.~\ref{fig:appendix_ss1m_vs_coco_text} gives a comparison of some text examples between SS1M and MS-COCO.

We explore using SSIM corpus to train our TIPCap under setting 3, where the image data is sampled from YFCC15M, identical to the experiments conducted on the MS-COCO.
After training, we evaluate TIPCap on MS-COCO and Flickr30K test datasets as shown in Table~\ref{tab:appendix_ss1m_s3} (middle 2 rows), the results indicate that TIPCap can be easily extended to larger datasets and also performs good zero-shot performance. Fig.~\ref{fig:appendix_examples} shows some examples of captions generated by TIPCap trained on SS1M under setting 3 (see ``SS1M to MS-COCO / Flickr30K''), which can correctly describe images, even with more descriptive details. 

% See Fig.~\ref{fig:appendix_examples_extend} and Fig.~\ref{fig:appendix_examples_extend_1} for more examples.

Another interesting results are also shown in Table~\ref{tab:appendix_ss1m_s3} (bottom 2 rows), we train our TIPCap on setting 1, where the required parameters $\vec{\mu}$ and $\Sigma$ are directly estimated from 1\% MS-COCO paired data. 
It can be seen that this paradigm achieves better performance, even though the distribution parameters are estimated from another dataset. 
We attribute this phenomenon to two reasons: 
1) As shown in Table~\ref{fig:appendix_ss1m_vs_coco_text}, the text corpus in SS1M have a good fluency and language accuracy; 
2) Comparing SS1M and MS-COCO, the styles of their text corpus are not exactly same but similar. Their text both consists of a main object and additional descriptive attribution like state and position.

\section{Conclusion}
In this paper, we {propose} a unified solution TIPCap with text-centric training data for image captioning, {which can} almost cover the vast majority of data configurations in real-world scenarios. 
In TIPCap,
the mapping module, which is driven by a multivariate Gaussian distribution and aims to mitigate the modality gap between image and text embedding.
Additionally, TIPCap can incorporate optional prompt information, which is proposed to improve generated captions.
Extensive experiments also demonstrate the effectiveness of our TIPCap.
We believe this study can give a new paradigm and benefit the community for image captioning.

% \newpage

\bibliographystyle{IEEEtran}
\bibliography{TIPCap}

\begin{thebibliography}{10}
\providecommand{\url}[1]{#1}
\csname url@samestyle\endcsname
\providecommand{\newblock}{\relax}
\providecommand{\bibinfo}[2]{#2}
\providecommand{\BIBentrySTDinterwordspacing}{\spaceskip=0pt\relax}
\providecommand{\BIBentryALTinterwordstretchfactor}{4}
\providecommand{\BIBentryALTinterwordspacing}{\spaceskip=\fontdimen2\font plus
\BIBentryALTinterwordstretchfactor\fontdimen3\font minus
  \fontdimen4\font\relax}
\providecommand{\BIBforeignlanguage}[2]{{%
\expandafter\ifx\csname l@#1\endcsname\relax
\typeout{** WARNING: IEEEtran.bst: No hyphenation pattern has been}%
\typeout{** loaded for the language `#1'. Using the pattern for}%
\typeout{** the default language instead.}%
\else
\language=\csname l@#1\endcsname
\fi
#2}}
\providecommand{\BIBdecl}{\relax}
\BIBdecl

\bibitem{show_and_tell}
O.~Vinyals, A.~Toshev, S.~Bengio, and D.~Erhan, ``Show and tell: {A} neural
  image caption generator,'' in \emph{Proc. IEEE Conf. Comput. Vis. Pattern
  Recognit.}, 2015, pp. 3156--3164.

\bibitem{show_attend_and_tell}
K.~Xu, J.~Ba, R.~Kiros, K.~Cho, A.~C. Courville, R.~Salakhutdinov, R.~S. Zemel,
  and Y.~Bengio, ``Show, attend and tell: Neural image caption generation with
  visual attention,'' in \emph{Proc. Int. Conf. Mach. Learn.}, 2015, pp.
  2048--2057.

\bibitem{updown}
P.~Anderson, X.~He, C.~Buehler, D.~Teney, M.~Johnson, S.~Gould, and L.~Zhang,
  ``Bottom-up and top-down attention for image captioning and visual question
  answering,'' in \emph{Proc. IEEE Conf. Comput. Vis. Pattern Recognit.}, 2018,
  pp. 6077--6086.

\bibitem{aoanet}
L.~Huang, W.~Wang, J.~Chen, and X.~Wei, ``Attention on attention for image
  captioning,'' in \emph{Proc. IEEE Int. Conf. Comput. Vis.}, 2019, pp.
  4633--4642.

\bibitem{Xtransformer}
Y.~Pan, T.~Yao, Y.~Li, and T.~Mei, ``X-linear attention networks for image
  captioning,'' in \emph{Proc. IEEE Conf. Comput. Vis. Pattern Recognit.},
  2020, pp. 10\,968--10\,977.

\bibitem{DLCT}
Y.~Luo, J.~Ji, X.~Sun, L.~Cao, Y.~Wu, F.~Huang, C.~Lin, and R.~Ji, ``Dual-level
  collaborative transformer for image captioning,'' in \emph{Proc. AAAI Conf.
  Artif. Intell.}, 2021, pp. 2286--2293.

\bibitem{PureT}
Y.~Wang, J.~Xu, and Y.~Sun, ``End-to-end transformer based model for image
  captioning,'' in \emph{Proc. AAAI Conf. Artif. Intell.}, 2022, pp.
  2585--2594.

\bibitem{DBLP:journals/tcsv/WuXSYM21}
\BIBentryALTinterwordspacing
L.~Wu, M.~Xu, L.~Sang, T.~Yao, and T.~Mei, ``Noise augmented double-stream
  graph convolutional networks for image captioning,'' \emph{{IEEE} Trans.
  Circuits Syst. Video Technol.}, vol.~31, no.~8, pp. 3118--3127, 2021.
  [Online]. Available: \url{https://doi.org/10.1109/TCSVT.2020.3036860}
\BIBentrySTDinterwordspacing

\bibitem{DBLP:journals/tcsv/JiangZH22}
\BIBentryALTinterwordspacing
W.~Jiang, W.~Zhou, and H.~Hu, ``Double-stream position learning transformer
  network for image captioning,'' \emph{{IEEE} Trans. Circuits Syst. Video
  Technol.}, vol.~32, no.~11, pp. 7706--7718, 2022. [Online]. Available:
  \url{https://doi.org/10.1109/TCSVT.2022.3181490}
\BIBentrySTDinterwordspacing

\bibitem{DBLP:journals/tcsv/CaoAZW22}
\BIBentryALTinterwordspacing
S.~Cao, G.~An, Z.~Zheng, and Z.~Wang, ``Vision-enhanced and consensus-aware
  transformer for image captioning,'' \emph{{IEEE} Trans. Circuits Syst. Video
  Technol.}, vol.~32, no.~10, pp. 7005--7018, 2022. [Online]. Available:
  \url{https://doi.org/10.1109/TCSVT.2022.3178844}
\BIBentrySTDinterwordspacing

\bibitem{DBLP:journals/tcsv/ZhangXDW23}
\BIBentryALTinterwordspacing
J.~Zhang, Y.~Xie, W.~Ding, and Z.~Wang, ``Cross on cross attention: Deep fusion
  transformer for image captioning,'' \emph{{IEEE} Trans. Circuits Syst. Video
  Technol.}, vol.~33, no.~8, pp. 4257--4268, 2023. [Online]. Available:
  \url{https://doi.org/10.1109/TCSVT.2023.3243725}
\BIBentrySTDinterwordspacing

\bibitem{mscoco}
T.~Lin, M.~Maire, S.~J. Belongie, J.~Hays, P.~Perona, D.~Ramanan,
  P.~Doll{\'{a}}r, and C.~L. Zitnick, ``Microsoft {COCO:} common objects in
  context,'' in \emph{Proc. Eur. Conf. Comput. Vis.}, 2014, pp. 740--755.

\bibitem{flickr30k}
P.~Young, A.~Lai, M.~Hodosh, and J.~Hockenmaier, ``From image descriptions to
  visual denotations: New similarity metrics for semantic inference over event
  descriptions,'' \emph{Trans. Assoc. Comput. Linguistics}, vol.~2, pp. 67--78,
  2014.

\bibitem{Feng_etal_unsupervised_IC}
Y.~Feng, L.~Ma, W.~Liu, and J.~Luo, ``Unsupervised image captioning,'' in
  \emph{Proc. IEEE Conf. Comput. Vis. Pattern Recognit.}, 2019, pp. 4125--4134.

\bibitem{DBLP:conf/iccv/Laina0N19}
I.~Laina, C.~Rupprecht, and N.~Navab, ``Towards unsupervised image captioning
  with shared multimodal embeddings,'' in \emph{Proc. IEEE Int. Conf. Comput.
  Vis.}, 2019, pp. 7413--7423.

\bibitem{BERT}
J.~Devlin, M.~Chang, K.~Lee, and K.~Toutanova, ``{BERT:} pre-training of deep
  bidirectional transformers for language understanding,'' in \emph{Proc. Conf.
  N. Am. Chapter Assoc. Comput. Linguistics: Hum. Lang. Technol.}, 2019, pp.
  4171--4186.

\bibitem{GPT2}
A.~Radford, J.~Wu, R.~Child, D.~Luan, D.~Amodei, and I.~Sutskever, ``Language
  models are unsupervised multitask learners,'' 2019.

\bibitem{T5}
C.~Raffel, N.~Shazeer, A.~Roberts, K.~Lee, S.~Narang, M.~Matena, Y.~Zhou,
  W.~Li, and P.~J. Liu, ``Exploring the limits of transfer learning with a
  unified text-to-text transformer,'' \emph{J. Mach. Learn. Res.}, vol.~21, pp.
  140:1--140:67, 2020.

\bibitem{CLIP}
A.~Radford, J.~W. Kim, C.~Hallacy, A.~Ramesh, G.~Goh, S.~Agarwal, G.~Sastry,
  A.~Askell, P.~Mishkin, J.~Clark, G.~Krueger, and I.~Sutskever, ``Learning
  transferable visual models from natural language supervision,'' in
  \emph{Proc. Int. Conf. Mach. Learn.}, 2021, pp. 8748--8763.

\bibitem{ALIGN}
C.~Jia, Y.~Yang, Y.~Xia, Y.~Chen, Z.~Parekh, H.~Pham, Q.~V. Le, Y.~Sung, Z.~Li,
  and T.~Duerig, ``Scaling up visual and vision-language representation
  learning with noisy text supervision,'' in \emph{Proc. Int. Conf. Mach.
  Learn.}, 2021, pp. 4904--4916.

\bibitem{BLIP}
J.~Li, D.~Li, C.~Xiong, and S.~C.~H. Hoi, ``{BLIP:} bootstrapping
  language-image pre-training for unified vision-language understanding and
  generation,'' in \emph{Proc. Int. Conf. Mach. Learn.}, 2022, pp.
  12\,888--12\,900.

\bibitem{cc12m}
S.~Changpinyo, P.~Sharma, N.~Ding, and R.~Soricut, ``Conceptual 12m: Pushing
  web-scale image-text pre-training to recognize long-tail visual concepts,''
  in \emph{Proc. IEEE Conf. Comput. Vis. Pattern Recognit.}, 2021, pp.
  3558--3568.

\bibitem{zerocap}
Y.~Tewel, Y.~Shalev, I.~Schwartz, and L.~Wolf, ``Zerocap: Zero-shot
  image-to-text generation for visual-semantic arithmetic,'' in \emph{Proc.
  IEEE Conf. Comput. Vis. Pattern Recognit.}, 2022, pp. 17\,897--17\,907.

\bibitem{magic}
Y.~Su, T.~Lan, Y.~Liu, F.~Liu, D.~Yogatama, Y.~Wang, L.~Kong, and N.~Collier,
  ``Language models can see: Plugging visual controls in text generation,''
  \emph{ArXiv}, vol. abs/2205.02655, 2022.

\bibitem{capdec}
D.~Nukrai, R.~Mokady, and A.~Globerson, ``Text-only training for image
  captioning using noise-injected {CLIP},'' in \emph{Proc. Conf. Empirical
  Methods Natural Lang. Process.}, 2022, pp. 4055--4063.

\bibitem{decap}
W.~Li, L.~Zhu, L.~Wen, and Y.~Yang, ``Decap: Decoding clip latents for
  zero-shot captioning via text-only training,'' in \emph{Proc. Int. Conf.
  Learn. Representations}, 2023.

\bibitem{close}
S.~Gu, C.~Clark, and A.~Kembhavi, ``I can't believe there's no images! learning
  visual tasks using only language data,'' \emph{ArXiv}, vol. abs/2211.09778,
  2022.

\bibitem{BLIP2}
J.~Li, D.~Li, S.~Savarese, and S.~C.~H. Hoi, ``{BLIP-2:} bootstrapping
  language-image pre-training with frozen image encoders and large language
  models,'' \emph{ArXiv}, vol. abs/2301.12597, 2023.

\bibitem{OFA}
P.~Wang, A.~Yang, R.~Men, J.~Lin, S.~Bai, Z.~Li, J.~Ma, C.~Zhou, J.~Zhou, and
  H.~Yang, ``{OFA:} unifying architectures, tasks, and modalities through a
  simple sequence-to-sequence learning framework,'' in \emph{Proc. Int. Conf.
  Mach. Learn.}, 2022, pp. 23\,318--23\,340.

\bibitem{SEEM}
X.~Zou, J.~Yang, H.~Zhang, F.~Li, L.~Li, J.~Gao, and Y.~J. Lee, ``Segment
  everything everywhere all at once,'' \emph{ArXiv}, vol. abs/2304.06718, 2023.

\bibitem{BLEU}
K.~Papineni, S.~Roukos, T.~Ward, and W.~Zhu, ``Bleu: a method for automatic
  evaluation of machine translation,'' in \emph{Proc. Annu. Meet. Assoc.
  Comput. Linguist.}, 2002, pp. 311--318.

\bibitem{DBLP:conf/emnlp/ChoMGBBSB14}
K.~Cho, B.~van Merrienboer, {\c{C}}.~G{\"{u}}l{\c{c}}ehre, D.~Bahdanau,
  F.~Bougares, H.~Schwenk, and Y.~Bengio, ``Learning phrase representations
  using {RNN} encoder-decoder for statistical machine translation,'' in
  \emph{Proc. Conf. Empirical Methods Natural Lang. Process.}, 2014, pp.
  1724--1734.

\bibitem{show_observe_and_tell}
H.~Chen, G.~Ding, Z.~Lin, S.~Zhao, and J.~Han, ``Show, observe and tell:
  Attribute-driven attention model for image captioning,'' in \emph{Proc. Int.
  Joint Conf. Artif. Intell.}, 2018, pp. 606--612.

\bibitem{DBLP:conf/cvpr/YouJWFL16}
Q.~You, H.~Jin, Z.~Wang, C.~Fang, and J.~Luo, ``Image captioning with semantic
  attention,'' in \emph{Proc. IEEE Conf. Comput. Vis. Pattern Recognit.}, 2016,
  pp. 4651--4659.

\bibitem{lstm}
S.~Hochreiter and J.~Schmidhuber, ``Long short-term memory,'' \emph{Neural
  Computation}, vol.~9, no.~8, pp. 1735--1780, 1997.

\bibitem{FasterRCNN}
S.~Ren, K.~He, R.~B. Girshick, and J.~Sun, ``Faster {R-CNN:} towards real-time
  object detection with region proposal networks,'' \emph{{IEEE} Trans. Pattern
  Anal. Mach. Intell.}, vol.~39, no.~6, pp. 1137--1149, 2017.

\bibitem{DBLP:conf/mm/DongLXX21}
X.~Dong, C.~Long, W.~Xu, and C.~Xiao, ``Dual graph convolutional networks with
  transformer and curriculum learning for image captioning,'' in \emph{Proc.
  ACM Int. Conf. Multimed.}, 2021, pp. 2615--2624.

\bibitem{DBLP:conf/mm/NieL0LL021}
W.~Nie, J.~Li, N.~Xu, A.~Liu, X.~Li, and Y.~Zhang, ``Triangle-reward
  reinforcement learning: {A} visual-linguistic semantic alignment for image
  captioning,'' in \emph{Proc. ACM Int. Conf. Multimed.}, 2021, pp. 4510--4518.

\bibitem{ORT}
S.~Herdade, A.~Kappeler, K.~Boakye, and J.~Soares, ``Image captioning:
  Transforming objects into words,'' in \emph{Proc. Adv. neural inf. proces.
  syst.}, 2019, pp. 11\,135--11\,145.

\bibitem{DBLP:journals/tcsv/YuLYH20}
\BIBentryALTinterwordspacing
J.~Yu, J.~Li, Z.~Yu, and Q.~Huang, ``Multimodal transformer with multi-view
  visual representation for image captioning,'' \emph{{IEEE} Trans. Circuits
  Syst. Video Technol.}, vol.~30, no.~12, pp. 4467--4480, 2020. [Online].
  Available: \url{https://doi.org/10.1109/TCSVT.2019.2947482}
\BIBentrySTDinterwordspacing

\bibitem{M2Transformer}
M.~Cornia, M.~Stefanini, L.~Baraldi, and R.~Cucchiara, ``Meshed-memory
  transformer for image captioning,'' in \emph{Proc. IEEE Conf. Comput. Vis.
  Pattern Recognit.}, 2020, pp. 10\,575--10\,584.

\bibitem{CLMs}
J.~Wang, Y.~Zhang, M.~Yan, J.~Zhang, and J.~Sang, ``Zero-shot image captioning
  by anchor-augmented vision-language space alignment,'' \emph{ArXiv}, vol.
  abs/2211.07275, 2022.

\bibitem{LXMERT}
H.~Tan and M.~Bansal, ``{LXMERT:} learning cross-modality encoder
  representations from transformers,'' in \emph{Proc. Conf. Empirical Methods
  Natural Lang. Process.}, 2019, pp. 5099--5110.

\bibitem{ViLBERT}
J.~Lu, D.~Batra, D.~Parikh, and S.~Lee, ``Vilbert: Pretraining task-agnostic
  visiolinguistic representations for vision-and-language tasks,'' in
  \emph{Proc. Adv. neural inf. proces. syst.}, 2019, pp. 13--23.

\bibitem{UNITER}
Y.~Chen, L.~Li, L.~Yu, A.~E. Kholy, F.~Ahmed, Z.~Gan, Y.~Cheng, and J.~Liu,
  ``{UNITER:} universal image-text representation learning,'' in \emph{Proc.
  Eur. Conf. Comput. Vis.}, 2020, pp. 104--120.

\bibitem{Oscar}
X.~Li, X.~Yin, C.~Li, P.~Zhang, X.~Hu, L.~Zhang, L.~Wang, H.~Hu, L.~Dong,
  F.~Wei, Y.~Choi, and J.~Gao, ``Oscar: Object-semantics aligned pre-training
  for vision-language tasks,'' in \emph{Proc. Eur. Conf. Comput. Vis.}, ser.
  Lecture Notes in Computer Science, vol. 12375, 2020, pp. 121--137.

\bibitem{ALBEF}
J.~Li, R.~R. Selvaraju, A.~Gotmare, S.~R. Joty, C.~Xiong, and S.~C. Hoi,
  ``Align before fuse: Vision and language representation learning with
  momentum distillation,'' in \emph{Proc. Adv. neural inf. proces. syst.},
  2021, pp. 9694--9705.

\bibitem{VAE}
D.~P. Kingma and M.~Welling, ``Auto-encoding variational bayes,'' in
  \emph{Proc. Int. Conf. Learn. Representations}, 2014.

\bibitem{instructGPT}
L.~Ouyang, J.~Wu, X.~Jiang, D.~Almeida, C.~L. Wainwright, P.~Mishkin, C.~Zhang,
  S.~Agarwal, K.~Slama, A.~Ray, J.~Schulman, J.~Hilton, F.~Kelton, L.~Miller,
  M.~Simens, A.~Askell, P.~Welinder, P.~F. Christiano, J.~Leike, and R.~Lowe,
  ``Training language models to follow instructions with human feedback,'' in
  \emph{Proc. Adv. neural inf. proces. syst.}, 2022.

\bibitem{DBLP:journals/pami/KarpathyF17}
A.~Karpathy and L.~Fei{-}Fei, ``Deep visual-semantic alignments for generating
  image descriptions,'' in \emph{Proc. IEEE Conf. Comput. Vis. Pattern
  Recognit.}, 2015, pp. 3128--3137.

\bibitem{yfcc100m}
B.~Thomee, D.~A. Shamma, G.~Friedland, B.~Elizalde, K.~Ni, D.~Poland, D.~Borth,
  and L.~Li, ``{YFCC100M:} the new data in multimedia research,'' \emph{Commun.
  {ACM}}, vol.~59, no.~2, pp. 64--73, 2016.

\bibitem{METEOR}
M.~J. Denkowski and A.~Lavie, ``Meteor universal: Language specific translation
  evaluation for any target language,'' in \emph{Proc. ACL Workshop Statistical
  Machine Translation}, 2014, pp. 376--380.

\bibitem{ROUGE}
C.-Y. Lin, ``{ROUGE}: A package for automatic evaluation of summaries,'' in
  \emph{Proc. ACL Workshop Text Summarization Branches Out}, 2004, pp. 74--81.

\bibitem{CIDEr}
R.~Vedantam, C.~L. Zitnick, and D.~Parikh, ``Cider: Consensus-based image
  description evaluation,'' in \emph{Proc. IEEE Conf. Comput. Vis. Pattern
  Recognit.}, 2015, pp. 4566--4575.

\bibitem{SPICE}
P.~Anderson, B.~Fernando, M.~Johnson, and S.~Gould, ``Spice: Semantic
  propositional image caption evaluation,'' in \emph{Proc. Eur. Conf. Comput.
  Vis.}, vol. 9909, 2016, pp. 382--398.

\bibitem{pytorch}
A.~Paszke, S.~Gross, F.~Massa, A.~Lerer, J.~Bradbury, G.~Chanan, T.~Killeen,
  Z.~Lin, N.~Gimelshein, L.~Antiga, A.~Desmaison, A.~K{\"{o}}pf, E.~Z. Yang,
  Z.~DeVito, M.~Raison, A.~Tejani, S.~Chilamkurthy, B.~Steiner, L.~Fang,
  J.~Bai, and S.~Chintala, ``Pytorch: An imperative style, high-performance
  deep learning library,'' in \emph{Proc. Adv. neural inf. proces. syst.},
  2019, pp. 8024--8035.

\bibitem{adamw}
I.~Loshchilov and F.~Hutter, ``Decoupled weight decay regularization,'' in
  \emph{Proc. Int. Conf. Learn. Representations}.\hskip 1em plus 0.5em minus
  0.4em\relax OpenReview.net, 2019.

\bibitem{uniVLP}
L.~Zhou, H.~Palangi, L.~Zhang, H.~Hu, J.~J. Corso, and J.~Gao, ``Unified
  vision-language pre-training for image captioning and {VQA},'' in \emph{Proc.
  AAAI Conf. Artif. Intell.}, 2020, pp. 13\,041--13\,049.

\bibitem{clipcap}
R.~Mokady, A.~Hertz, and A.~H. Bermano, ``Clipcap: {CLIP} prefix for image
  captioning,'' \emph{ArXiv}, vol. abs/2111.09734, 2021.

\bibitem{nocaps}
H.~Agrawal, P.~Anderson, K.~Desai, Y.~Wang, X.~Chen, R.~Jain, M.~Johnson,
  D.~Batra, D.~Parikh, and S.~Lee, ``nocaps: novel object captioning at
  scale,'' in \emph{Proc. IEEE Int. Conf. Comput. Vis.}, 2019, pp. 8947--8956.

\end{thebibliography}

\vfill

\end{document}